\documentclass[letterpaper, 10 pt, conference]{ieeeconf}  % Comment this line out if you need a4paper
\IEEEoverridecommandlockouts
\usepackage{cite}
\usepackage{amsmath,amssymb,amsfonts}
\usepackage{algorithmic}
\usepackage{graphicx}
\usepackage[hyphens,spaces,obeyspaces]{url}
\usepackage{hyperref}
\usepackage{todonotes}
\usepackage{textcomp}
\usepackage{xcolor}
\usepackage{bm}
\usepackage{threeparttable}
\usepackage{siunitx}
\usepackage{multirow}
\def\BibTeX{{\rm B\kern-.05em{\sc i\kern-.025em b}\kern-.08em
    T\kern-.1667em\lower.7ex\hbox{E}\kern-.125emX}}

\hypersetup{
	colorlinks=true,
	linkcolor=blue,
	filecolor=magenta,      
	urlcolor=blue,
}

\begin{document}

%\title{\LARGE \bf The Atlas Benchmark: an Automated Evaluation Toolkit \\ for Human Motion Prediction Algorithms
\title{\LARGE \bf The Atlas Benchmark: an Automated Evaluation Framework \\ for Human Motion Prediction
}

\author{Andrey Rudenko$^{1,3}$, Luigi Palmieri$^{1}$, Wanting Huang$^{1,2}$, Achim J. Lilienthal$^{3}$ and Kai O. Arras$^{1}$% <-this % stops a space
\thanks{This work was supported by the European Union's Horizon 2020 research and innovation program under grant agreement No. 101017274 (DARKO)}% <-this % stops a space
\thanks{$^{1}$Robert Bosch GmbH, Corporate Research, Stuttgart, Germany {\tt\small \{andrey.rudenko, luigi.palmieri, kaioliver.arras\}@de.bosch.com}}%
\thanks{$^{2}$TU M\"unchen, Germany
{\tt\small huangwt1994@163.com}}%
\thanks{$^{3}$Mobile Robotics and Olfaction Lab,
	\"Orebro University, Sweden
{\tt\small achim.lilienthal@oru.se}}%
}

\maketitle

\begin{abstract}
	%%% Previous version, koa
	%%%Human motion trajectory prediction, an essential task for autonomous systems in many domains, has been on the rise in the recent years. With a multitude of new methods proposed by different communities, the lack of standardized benchmarks and objective comparison between them has been a limitation for assessing progress and guiding further research. The few prior art benchmarks do not cover the full spectrum of important experiments and do not sufficiently include necessary contextual cues about the moving people and the environment. In this paper we present the Atlas benchmark which is designed for evaluation and comparison in automated reproducible experiments with a systematic variation of the key motion prediction parameters. The Atlas benchmark offers tools, such as metrics, data preparation and filtering, method calibration and visualization, and includes a large variety of heterogeneous datasets, representing usual human motion behaviors in different places and cultures. Using Atlas we study several popular model- and learning-based methods and discuss their strengths and limitations in terms of prediction accuracy, transfer to new environments, and robustness to perception noise and limited observations. 
	
	%%% koa version
	Human motion trajectory prediction, an essential task for autonomous systems in many domains, has been on the rise in recent years. With a multitude of new methods proposed by different communities, the lack of standardized benchmarks and objective comparisons is increasingly becoming  %identified to be
	a major limitation to assess progress and guide further research. Existing benchmarks are limited in their scope and flexibility to conduct relevant experiments and to account for contextual cues of agents and environments. In this paper we present Atlas, a benchmark to systematically evaluate human motion trajectory prediction algorithms in a unified framework. Atlas offers data preprocessing functions, hyperparameter optimization, comes with popular datasets and has the flexibility to setup and conduct underexplored yet relevant experiments to analyze a method’s accuracy and robustness. In an example application of Atlas, we compare five popular model- and learning-based predictors and find that, when properly applied, early physics-based approaches are still remarkably competitive. Such results confirm the necessity of benchmarks like Atlas.

	%to overcome several limitations of existing benchmarking. 
	%sustaining the enduring development of better algorithms.
\end{abstract}

%\begin{IEEEkeywords}
%component, formatting, style, styling, insert
%\end{IEEEkeywords}

%\section{Introduction}
%Human motion trajectory prediction is an essential task in robotics. A mobile robot operating in social spaces should be able to predict human motion in order to plan safe and efficient paths towards its goal. In the past few years, more methods have been proposed for human trajectory prediction responding to the emerging applications of artificial intelligence like autonomous cars and service robots \cite{rudenko2020human}.

%Existing benchmarks like TrajNet++ compare the performance of different methods used in datasets and do prediction based on chosen frames. However, the visualizer of this benchmark can only show the chosen single pedestrian in frames, so users can not intuitively observe the interactions among pedestrians. Besides, our benchmark not only offers more tunable parameters like prediction horizons, whether imports semantic maps and supposed destination, but also provides the choice to add noise to datasets for checking the robustness of algorithms and probabilistic prediction results.

\section{Introduction}

Benchmarking motion prediction algorithms is a challenging task. The evaluation outcome can be affected by various factors such as data, parameters, hyperparameters and experiment design. Elaborate and carefully designed experiments are necessary to expose specific abilities or limitations of a method, in particular for complex learning approaches. Influencing factors are, for example, the observation period, i.e. the duration that agents need to be seen to allow for accurate prediction of their motion, or the exact procedure how to set up a testing scenario from sequences of raw person detections. Even when evaluating a simple constant velocity motion model with the same dataset, metrics and prediction horizon, the evaluation results may still vary as reported in  \cite{alahi2016social} and \cite{scholler2019simpler} due to differences in testing scenario generation and data pre-processing. %The problem is further exemplified when evaluating on datasets, prohibitively large and with small amounts of non-trivial motion and social interactions, to justify using the entire dataset for evaluation
The limitations of the protocols commonly used to evaluate new prediction methods have been pointed out by several authors \cite{becker2018red,scholler2019simpler,rudenko2020human}.

\begin{figure}[t] 
	\centering
	\includegraphics[width=0.48\textwidth]{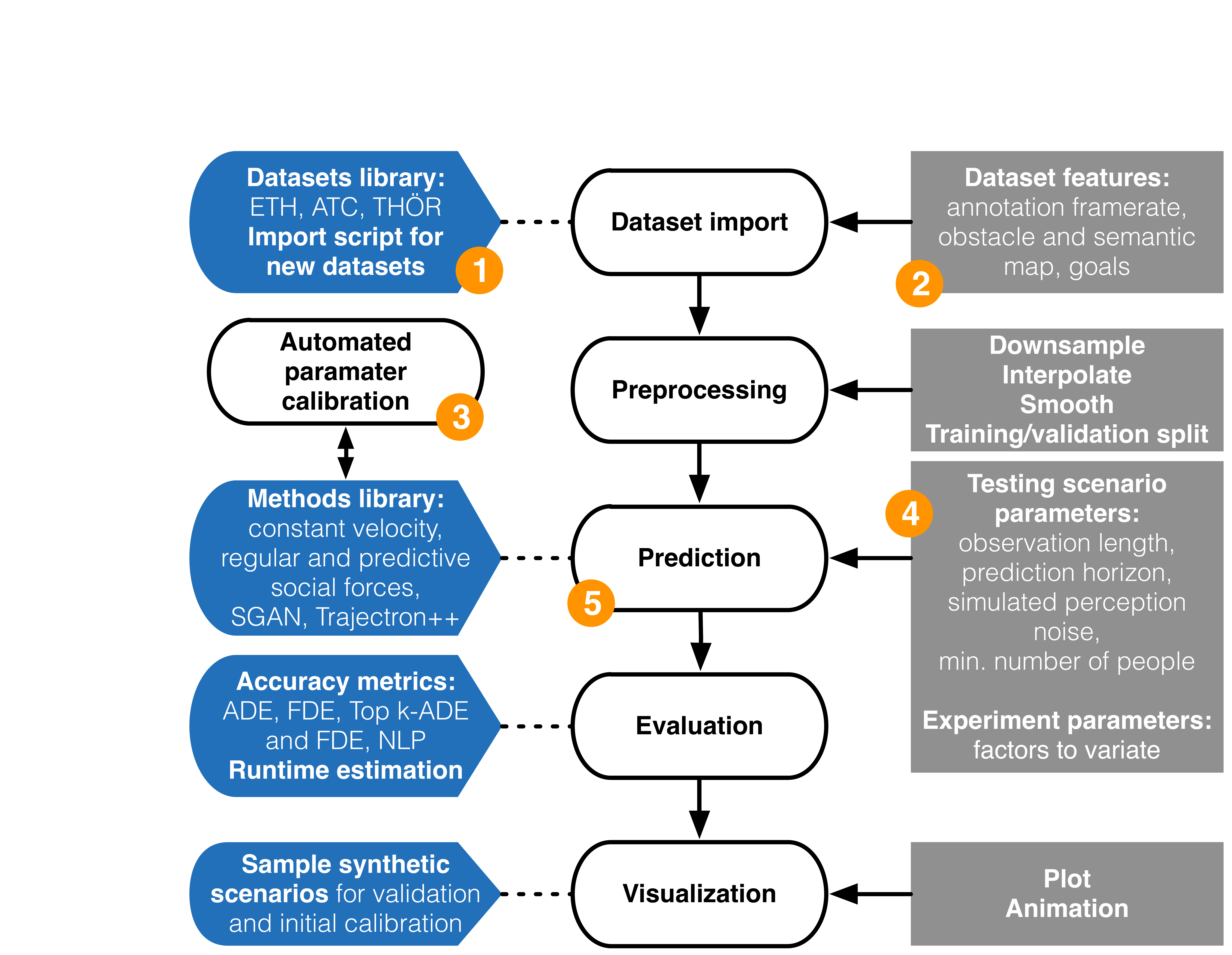}
	\caption{Atlas benchmark overview: (1) Supported import of new datasets (labeled detection streams), (2) Support for contextual cues in the environment, (3) Automated calibration of prediction hyperparameters, (4) Automated parametrized scenario extraction, (5) Direct interface to the prediction methods.} 
	\label{flow}
\end{figure}
% \todo[inline]{LP: What about changing this figure a bit?? is this the same of the workshop submission? Also extending the caption can be beneficial.}
In this paper we present the Atlas benchmark as a first step towards automated benchmarking of motion prediction methods in a unified framework with systematic variation of parameters. Atlas includes heterogeneous datasets of human motion trajectories, is capable of automatically extracting testing scenarios, and can deal with varying, missing and noisy agent detections using data interpolation, downsampling and smoothing\footnote{The Atlas benchmark will be available at \url{https://github.com/boschresearch/the-atlas-benchmark}}. Compared to prior art such as TrajNet++ \cite{kothari2020human}, it offers several tunable parameters like the observation period and prediction horizon, is able to import semantic maps and other relevant information such as goal positions in the map, allows to evaluate probabilistic prediction results and to conduct robustness experiments with simulated perception noise. Due to those features, our benchmark works with both short- and long-term predictors. Unlike TrajNet++, it is especially suited for studying how prediction parameters influence the results, in contrast to fixing the main parameters to produce the ranking scores in a specific \emph{challenge}. Furthermore, our benchmark has a direct interface to the hyperparameter estimation framework SMAC3 \cite{smac-2017} to calibrate a predictor on a specific dataset. This feature is particularly useful for model-based predictors, which, as we will show in the experiments, can perform still very well compared to recent learning-based ones.

We showcase Atlas by evaluating several popular model- and learning-based methods \cite{helbing1995social, karamouzas2009predictive, Gupta2018SocialGAN, salzmann2020trajectronpp} in terms of their prediction accuracy, ability to predict in new environments, and their robustness to perception noise and limited observations.

The paper is structured as follows: in Sec.~\ref{sec:background} we define the problem of benchmarking motion prediction and review prior art, and in Sec.~\ref{sec:atlas} we introduce our benchmark. The methods, evaluation setup and results are presented in Sec.~\ref{sec:experiments}, and Sec.~\ref{sec:conclusion} concludes the paper.

\section{Background}
\label{sec:background}

A trajectory prediction method aims to estimate a probability distribution over future positions of a moving agent within a certain time horizon. Typically, a motion predictor uses as input the agent's current or past motion states, possibly augmented by the current or past states of the environment. The environment is represented by the states of other moving agents, a topometric map of static obstacles, and possibly semantic information associated to parts, locations or objects of the map.

For the evaluation of a motion predictor we consider the following elements: datasets (popular examples include \cite{pellegrini2009you,lerner2007crowds,majecka2009statistical,brscic2013person,robicquet2016learning,rudenko2020thor}), the testing scenario extraction strategy and the evaluation metrics. As testing scenario extraction we denote the conversion of the continuous flow of (agent) detections, where past detections between consecutive frames form the observation history of length $O_s \in \mathbb{R}^+$ seconds (or $O_p \in \mathbb{Z}^+$ positions), and future agent states within horizon $T_s \in \mathbb{R}^+$ seconds (equivalent to $T_p\in \mathbb{Z}^+$ positions) form the predictions to be compared to the ground truth (GT). The metrics used to this end include geometric and probabilistic distance estimates between predicted and GT positions \cite{rudenko2020human}.

% This outlines the main parameters of the evaluation: the dataset, the testing scenario extraction strategy, $O_p$ and $T_p$ durations, and finally the metrics.

%%% koa: rewritten proposal below. Earlier version:
%The evaluation strategy, proposed by Alahi et al. \cite{alahi2016social} and formalized in the historically first \emph{TrajNet} benchmark for motion prediction \cite{sadeghiankosaraju2018trajnet}, has been adopted by many authors \cite{sadeghian2018sophie,huang2019stgat,kosaraju2019social,nikhil2018convolutional,huynh2019trajectory,xue2018ss,zhao2019multi,zhang2019sr,amirian2019social}. It uses the ETH and UCY datasets with fixed $O_s=3.2$ \SI{}{\second} and $T_s=4.8$ \SI{}{\second} and the ADE and FDE geometric metrics. TrajNet does not include variability in the main parameters $O_s$ and $T_s$, obstacles in the environment and any notion of prediction uncertainty or robustness.

Alahi et al. were the first to propose a benchmark for human trajectory prediction, called \emph{TrajNet} \cite{sadeghiankosaraju2018trajnet}. TrajNet has been used by many authors \cite{sadeghian2018sophie,huang2019stgat,kosaraju2019social,nikhil2018convolutional,huynh2019trajectory,xue2018ss,zhao2019multi,zhang2019sr,amirian2019social} and implements the evaluation strategy in \cite{alahi2016social}: it uses the ETH and UCY datasets with fixed $O_s=3.2$ \SI{}{\second} and $T_s=4.8$ \SI{}{\second} and the geometric metrics ADE and FDE. TrajNet does not include variability in the main parameters $O_s$ and $T_s$, obstacles in the environment, nor any notion of prediction uncertainty or robustness.

\emph{TrajNet++} \cite{kothari2020human} improves TrajNet by including additional  
%An improved \emph{TrajNet++} benchmark \cite{kothari2020human} uses several further 
datasets and it can further be extended with new ones (stored in json format). The benchmark focuses on evaluating agent interaction modeling approaches, and offers categorization of scenarios into classes of motion. It includes the possibility to predict several discrete positions for each pedestrian in each step, but does not support other probability distribution representations. The main limitation here, however, are the rigidly defined testing parameters, which restrict the evaluation to fixed parameters $O_s=3.2$ \SI{}{\second} and $T_s=4.8$ \SI{}{\second}. Furthermore, the scenario extraction strategy only guarantees that in each scenario \emph{one} target pedestrian has a complete track of requested $O_p+T_p$ consecutive positions. This contradicts the assumption, commonly made by many authors, that the history tracks for \emph{all} pedestrians are available at the time of prediction \cite{bartoli2017context,fernando2018soft,amirian2019social,tao2020dynamic}. Methodologically, considering scenarios with full observation tracks allows studying the effects of having limited (as well as abundant) observations for all agents. This approach allows isolating the prediction error caused by insufficient observations of the surrounding agents, even when observing the target one sufficiently long. Finally, TrajNet++ does not support obstacle and semantic information about the environment.

Based on these insights, we developed the \emph{Atlas benchmark} with an automated procedure to extract testing scenarios from datasets with flexible $O_p$ and $T_p$ parameters. Atlas accepts occupancy and semantic maps as input, supports various forms of parametric and non-parametric uncertainty representation, and includes robustness experiments with added noise to the observed trajectories.
% \todo[inline]{LP: Shall we include here also the fact that we show how to perform automatic extraction of best parameters with SMAC?}
%We outline the properties of each benchmark in Table~\ref{tab:benchmark_comparison}.

%%% Original version, see below
%Other notable advances in motion prediction benchmarking include the challenges in the automated driving workshops at NeurIPS 2019\footnote{\url{https://ml4ad.github.io/2019/}} and CVPR 2020\footnote{\url{http://cvpr2020.wad.vision/}} (``Workshop on Autonomous Driving'') based on the Argo dataset. NuScenes has a challenge based on their dataset\footnote{\url{https://www.nuscenes.org/}}. Recently, also the Interaction dataset featured the Interpret challenge at NeurIPS 2020\footnote{\url{http://challenge.interaction-dataset.com/prediction-challenge}}. These challenges concentrate on the specific automated driving use-case and one specific dataset. Furthermore, in the robotics domain, Hug~et~al.~\cite{hug2020short} proposed a Single Trajectory Sanity Check Benchmark, currently under construction\footnote{\url{https://stsc-benchmark.github.io/}}, which shares some aspects with Atlas in studying the effects of trajectory pre-processing.

%%% Version by koa
Other recent advances in motion prediction benchmarking include the challenges in the automated driving workshops at NeurIPS 2019\footnote{\url{https://ml4ad.github.io/2019/}}, NeurIPS 2020\footnote{\url{http://challenge.interaction-dataset.com/prediction-challenge}} and CVPR
2020\footnote{\url{http://cvpr2020.wad.vision/}} based on the Argo and Interaction datasets. NuScenes has an own challenge based on their dataset\footnote{\url{https://www.nuscenes.org/}}. These challenges concentrate on  automated driving only and on specific datasets. In robotics, Hug~et~al.~\cite{hug2020short} proposed a Single Trajectory Sanity Check Benchmark, currently under construction\footnote{\url{https://stsc-benchmark.github.io/}}. While these efforts share some aspects with Atlas, e.g., that they allow to study the effects of data pre-processing, they are limited in scope and flexibility, focusing on a single use-case, a single dataset and the generation of leaderboards for which, for example, main parameters remain fixed.

%\begin{figure}[t] 
%	\centering
%	\includegraphics[width=0.999\columnwidth]{fig/atc_test_scenario.pdf}
%	\caption[Example of data pre-processing in the Atlas benchmark]{Data processing example in the ATC dataset. Horizontal tracks in {\bf black} show available trajectory data (i.e. detections) for each frame on the x axis. Missing positions between detections, which can be interpolated, are shown with {\bf red} lines. Vertical lines in {\bf blue} show the observed track and prediction horizon. In this testing scenario, defined by the observed tracks, predictions are made for 14 people, but prediction for person 7 is excluded from evaluation as no ground truth positions are available.}
%	\label{datarecons}
%\end{figure}

\section{The Atlas Benchmark}
\label{sec:atlas}

Atlas includes five main elements: data import, preprocessing, the actual prediction phase, evaluation and visualization, see Fig.~\ref{flow}.
%Its design is a result of a deep analysis of the currently available datasets and desired tools for benchmarking of human motion prediction algorithms.
%%% koa: I don't get this sentence
%%%By explicitly interfacing the prediction module and scripting the experiments, our benchmark is suited for flexible and highly automated assessment of the motion prediction algorithms.
This design allows to interface and parametrize different prediction algorithms for a flexible and highly automated evaluation and analysis.

As first step, the datasets and, if available, information on the environment such as goals, obstacles, and semantics are imported into the benchmark.
%, which may enhance the prediction performance in some scenes.
Next, the raw data is preprocessed with downsampling to a user-defined frequency, misdetection interpolation and trajectory smoothing.
%, and reconstructed in the form of Fig.~\ref{datarecons}. 
Once the dataset is ready, we extract the testing scenarios with the user-specified observation and prediction lengths, and the minimum number of observed people.
%In this way when we give the start frame, end frame and prediction steps, we can obtain predictions. 
The observed past trajectories of all people in the testing scenario, along with environment data, are explicitly interfaced as input to the prediction algorithm. The returned predictions are evaluated against the ground truth using several metrics. Finally, the prediction results can be visualized with plots or animations. Meta-parameters to control the data processing and benchmark setup are stored in separate yaml files, and the benchmark is accessed via Jupyter notebooks.

\begin{figure}[t] 
	\centering
	\includegraphics[width=0.452\textwidth]{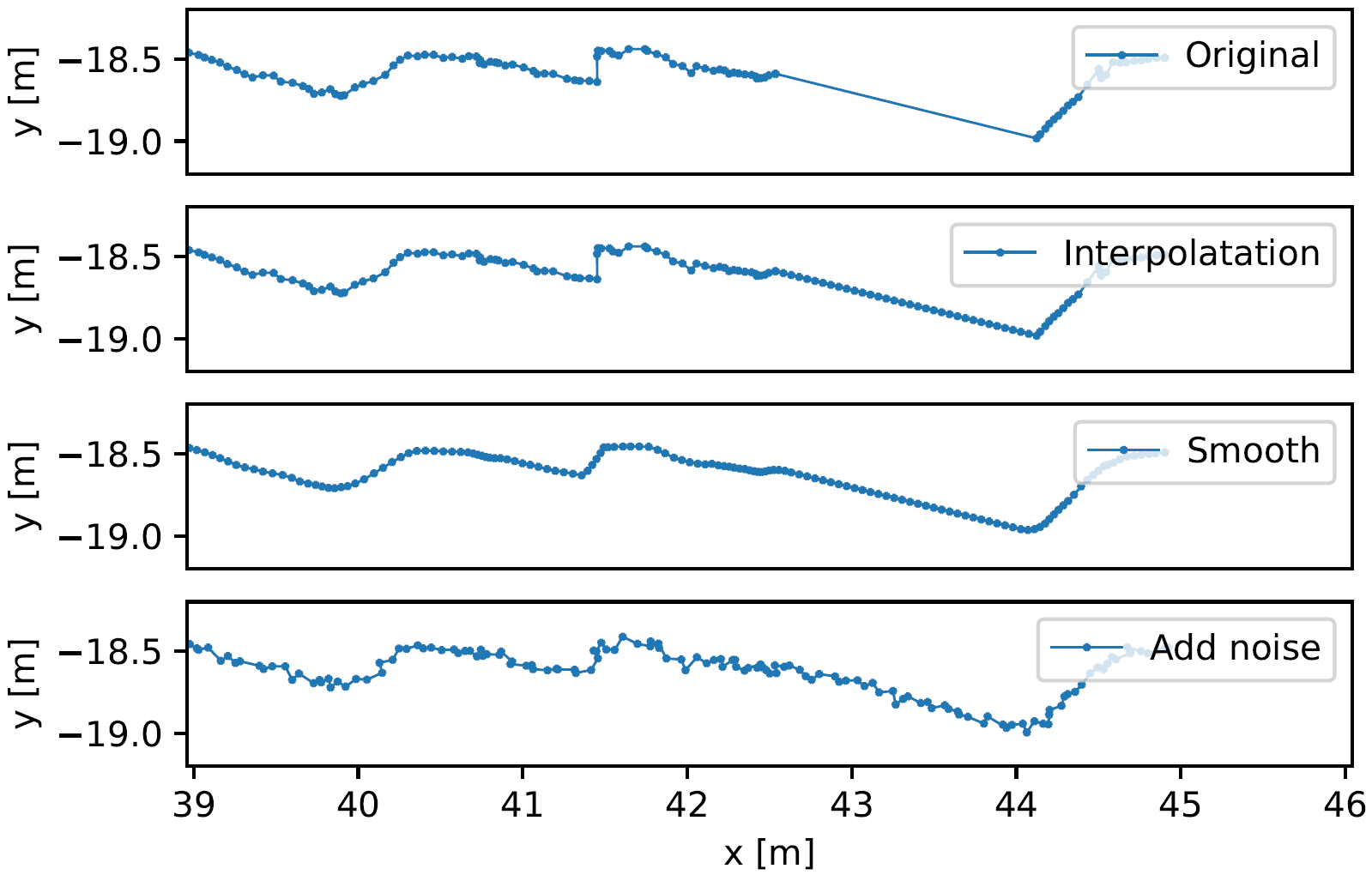}
	%\vspace{-8pt}
	\caption{Data preprocessing in Atlas. Example trajectory from the ATC dataset with noise and missing detections (\emph{original}, on top). The same trajectory is shown with misdetection interpolation, smoothing and adding a controlled amount of noise.}
	%\vspace{-5pt}
	\label{preprocess}
\end{figure}

\begin{figure*}[t]
	\centering	
	\includegraphics[width=0.432\textwidth]{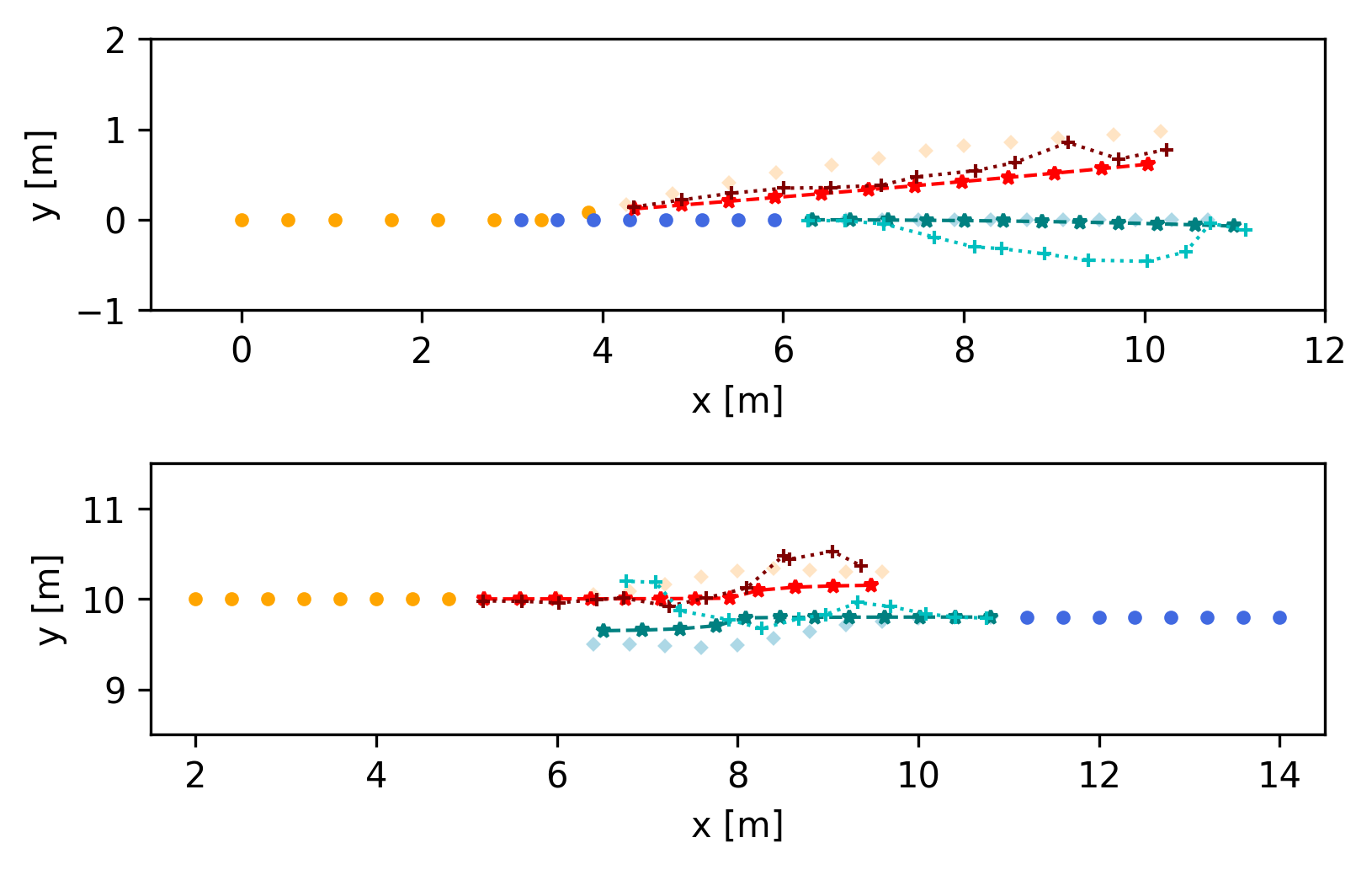}
	\includegraphics[width=0.558\textwidth]{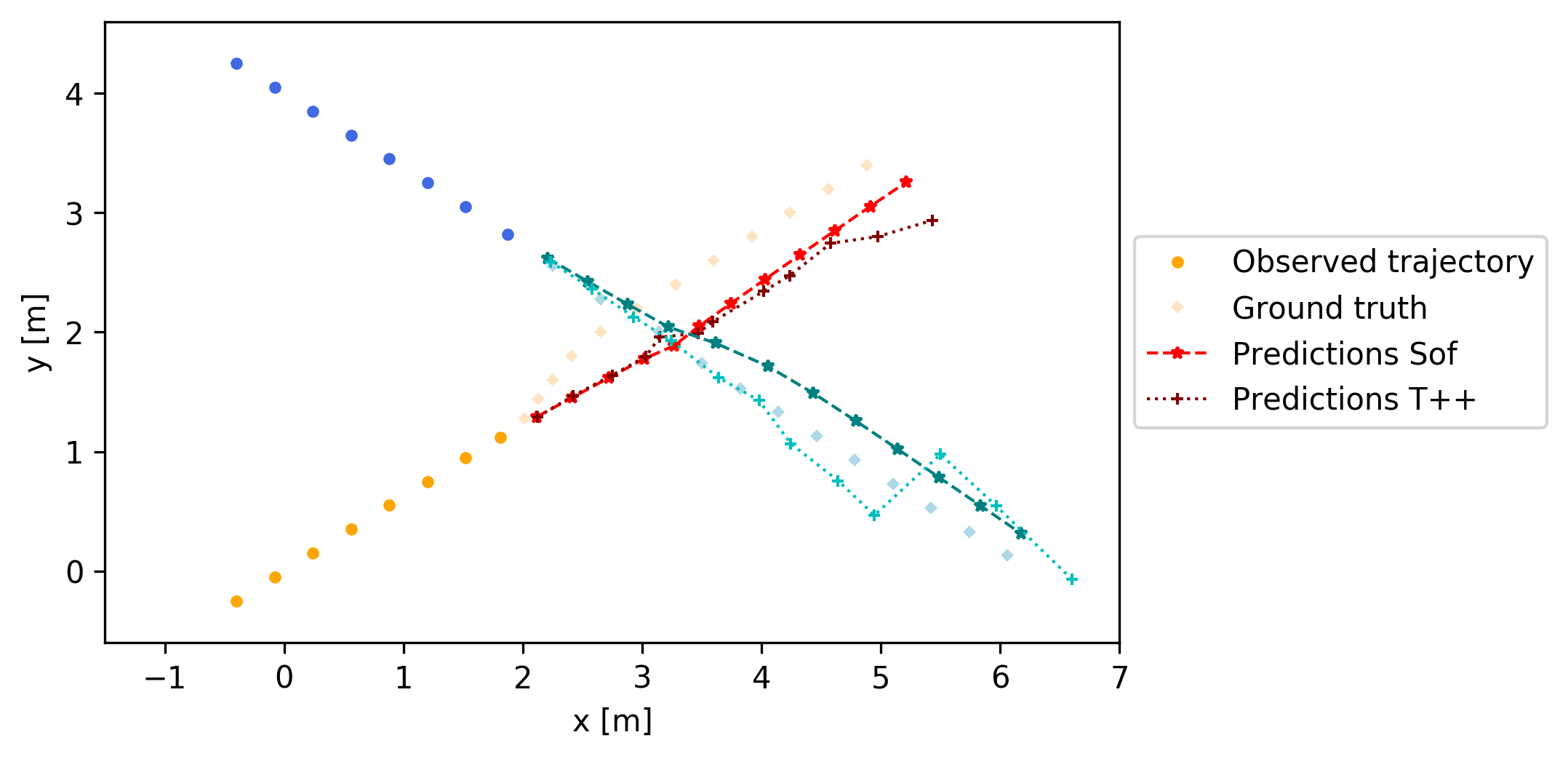}
	%\vspace{-8pt}
	\caption{Synthetic test scenarios in Atlas: \emph{chasing} (top left), \emph{opposing} (bottom left), and \emph{crossing} (right). The {\bf blue} and {\bf orange} dots show the observations of two simulated agents, the {\bf dashed} lines show the social force predictions \cite{helbing1995social}, and the {\bf dotted} lines the Trajectron++ results \cite{salzmann2020trajectronpp} as two example methods. These scenarios are used for initial calibration and inspection of the methods.}
	%\vspace{-5pt}
	\label{testing}
\end{figure*}

%In the following sections we describe each step of the benchmark in more detail.
\subsection{Datasets}
\label{sec:atlas:datasets}

Benchmark users can import any dataset in the specific json file format defined by TrajNet++ \cite{kothari2020human}, which includes for each detection the time stamp, person id and position. The json dataset format also supports obstacles and semantic grid maps \cite{rudenko2020semapp}, as well as goals (i.e. possible destinations of people) in the environment, which may influence the possible destinations of people.

Our benchmark currently includes the following three datasets:
\begin{itemize}
	\item[\emph{i)}] \textbf{ETH} \cite{pellegrini2009you}: This dataset contains people detections from video data recorded outdoors in the ETH campus in two locations: ETH and HOTEL.
	\item[\emph{ii)}] \textbf{ATC} \cite{brscic2013person}: Recorded in a shopping mall in Japan, the dataset covers a large indoor environment with densely crowded scenes.
	\item[\emph{iii)}] \textbf{TH\"OR} \cite{rudenko2020thor}: This dataset captures human motion in a room with static obstacles. It includes a setup with one obstacle (denoted as TH\"OR1, see Fig.~\ref{fig:atlas:predictions:thor1}) and with three obstacles (which we call TH\"OR3, see Fig.~\ref{fig:atlas:predictions:thor3}).
\end{itemize}

These three datasets come from different countries and were recorded in different environments, which increases the diversity of the scenarios and allows comparing prediction methods on different social and cultural contexts. For an in-depth analysis of the datasets we refer the reader to \cite{rudenko2020thor} and \cite{rudenko2020human}. In addition, we provide a possibility to import any dataset of labeled detections, as defined in Sec.~\ref{sec:background}.%, into the Atlas benchmark.
%and will make the evaluation results more convincing.
\subsection{Preprocessing}
\label{sec:atlas:preprocessing}
%The raw datasets are only recorded position information marked by pedestiran's identity and single frame number, so we need to reconstruct it to pedestrians' trajectories in chosen frame period, like the form of Fig.~\ref{datare}.The end user can freely choose the start frame and end frame, which determine the observed history length of trajectory.
Raw datasets often include noise and annotation artifacts (e.g. missing detections) \cite{rudenko2020thor}. Hence, our benchmark offers interpolation and smoothing in the preprocessing step. In addition, as a way to evaluate robustness of prediction algorithms, white Gaussian noise may be added to each detection. Fig.~\ref{preprocess} shows the preprocessing steps applied to an example trajectory in the ATC dataset. After detecting the missing frames in the original trajectory based on the average annotation frequency, we interpolate the points linearly in the missing part of the trajectory. Then, a moving average filter is used to smooth the noise.
%The window length of the filter is a parameter in our benchmark, in Fig.~\ref{fig:atlas:preprocess} it equal to 5.
Finally, random noise distributed as $\mathcal{N}(0,\sigma^2)$, where $\sigma$, can be added to each detection.

After preprocessing, we generate the testing scenarios with the observation length $O_p$ and ground truth for the following $T_p$ frames. As prediction performance may strongly depend on the observation length (in particular for intention estimation or when person detections are noisy), it is critical that all people in the testing scenario are observed in each of the $O_p$ frames. A testing scenario, along with the environment information, is then passed to the prediction step.

\subsection{Prediction}
\label{sec:atlas:prediction}
Our benchmark offers a direct interface to the prediction module, which is called at this step for the given testing scenario. This allows automated evaluation with a systematic variation of parameters, defined at the previous steps. For optimizing the hyperparatemers of the prediction methods, such as \cite{helbing1995social,karamouzas2009predictive, zanlungo2011social, kim2015brvo,farina2017walking}, Atlas includes an interface to the SMAC3 optimizer \cite{smac-2017}.
% \todo[inline]{LP: As stated above, we could highlight this SMAC3 integration as a contribution of the approach -- when compared to other baselines.}

Prior to benchmarking the prediction model on real data, the users can first validate their methods with several synthetic scenarios, which model fundamental interactions between people and the environment, e.g. individuals and groups walking in the opposite directions, crossing paths and navigating around hindrances (see several examples in Fig.~\ref{testing}). For instance, Fig.~\ref{testing} (bottom left) shows two people walking on a collision course towards each other. Their velocities are 1 \SI{}{\metre\per\second} and the initial displacement in the $y$ axis is 0.2 \si{\metre}. The frame frequency is \SI{2.5}{\Hz} and $O_p$ is 8 (frames). %Fig.~\ref{testing} also includes example predictions made by two popular social force-based models \cite{helbing1995social,zanlungo2011social}.
%From this figure we can observe Zanlungo's model will turn around earlier than Helbing's model so it has more foresight.
%\todo{We could include some showcase picture of those}.
%ANDREY: Removed this passage, it is described in the previous section.
%Meanwhile, there are two benchmark parameters can be tuned, observation length and prediction horizon, which determines the length of observed past trajectory, and the number of prediction steps.

Our benchmark supports various forms of parametric and non-parametric uncertainty representations for the prediction results. Non-parametric particle-based uncertain predictions are represented with a set of $K$ discrete sampled positions for each timestep. Alternatively, the results can be encoded as probabilities in 2D grid-map states, separately for each person in each timestep. Parametric uncertainty can be represented with a mixture of Gaussians: each individual mode of motion $z_i \in [z_1,\dots,z_K]$ is given as a sequence of Gaussian distributions
$\left( \mathcal{N}(\mathbf{\mu}_1, \mathbf{\Sigma}_1), \ldots, \mathcal{N}(\mathbf{\mu}_{T_p}, \mathbf{\Sigma}_{T_p}) \right)$, 
and the full predicted probability for timestep $i$ is $\sum_i \pi_i \mathcal{N}(\mathbf{\mu}_i, \mathbf{\Sigma}_i), i \in [1,\dots,T_p]$, where $\pi_i$ are the mixture weights. These options allow the evaluation of most existing prediction algorithms.
% \todo[inline]{LP: Shall we be more pedantic about notation here? How do we defined the GMMs? a name for the number of samples?}

%\subsubsection{Synthetic testing scenarios}
%\subsubsection{Benchmark parameters}There are two parameters, 
%\subsubsection{Method parameters}These parameters  can be useful for some prediction models, such as goal and obstacle. For example, for social force model, known goal information can help to calclate intended velocity $v_0$, so end users can choose whether to use goal in prediction. Hence we can also call them \textit{Boolen Parameter}. In addition, when calculating initial intended velocity $v_0$, there are two filters offered, gaussian and linear average filtering. Gaussian filtering gives the velocity of latest time step highest weight while linear average filtering gives the velocity of all history time step same weight.
\subsection{Evaluation}
\label{sec:atlas:evaluation}
The Atlas benchmark supports geometric and probabilistic metrics, as defined in \cite{rudenko2020human}. Geometric metrics include the \textit{Average Displacement Error} (ADE), which describes the error between points of predicted trajectories and the ground truth at the same timestep, and the \textit{Final Displacement Error} (FDE), which computes the error at the last prediction step. Probabilistic metrics include the \emph{Negative Log-Probability} (NLP), which computes the average probability of the ground truth position under the predicted distribution for the corresponding frame, and \emph{Top-k ADE and FDE}, which compute the displacements between the ground truth position and the closest of the $K$ samples from the predicted distribution.

\subsection{Experiments}
\label{sec:atlas:experiments}

Building on the datasets, pre-processing steps and metrics described above, the benchmark enables researchers to set up and conduct several experiments to study prediction performance under varying conditions. Such experiments are not only key for researchers to better understand the algorithm or model at hand, e.g. during ablation studies, but also for practitioners to evaluate a predictor within a system with adjacent up- and downstream tasks and real-world deployments. %, in contrast to a limited insight contained in a single benchmark score.
%Due to the automated nature of our benchmark, the experiments are scripted with all parameters available externally in a yaml file.

\subsubsection{Prediction accuracy conditioned on parameters}

$O_s$ and $T_s$ are among the main factors, associated with predicting motion. The accuracy naturally degrades for further time instances, while longer observations may improve it overall. In Atlas it is possible to measure the accuracy of prediction conditioned on these two main parameters. Further accuracy breakdown is possible by conditioning the measured values on the number of people in the scenario.

\subsubsection{Transfer experiments}

A crucial part of evaluating a prediction method is to analyze its generalization ability to new environments not included in the training data. Such experiments are most often overlooked in related work. In Atlas it is possible to script hyperparameter optimization in one dataset, and evaluate the method in another. In the future we plan to extend this functionality for training models.

\subsubsection{Robustness experiments}

For a system to work in the real world, a predictor must be robust against imperfection in perception such as noisy agent position observations.
 One possible way to quantify robustness, implemented in Atlas, is by measuring accuracy on the testing scenarios, after artificially adding increasing amounts of white Gaussian noise. % to the initially noise-free data.

%\begin{figure}[t] 
%	\centering
%	\includegraphics[scale=0.6]{fig/gmm.pdf}
%	\caption{A testing scenario from the ATC dataset. This figure shows the probability distributions over the final positions of the observed people. Each person's past track and ground truth future motion is shown with a different color. Ellipses show the \emph{analytical} and points show the \emph{particle-based} probability distribution over the final position.}
%	\label{gmm}
%\end{figure}

\section{Example Evaluation}
\label{sec:experiments}

With the Atlas benchmark described above, we now demonstrate its usage in an example evaluation. To this end, we conducted experiments to study and compare the performance of a small range of popular methods for human motion prediction, from simple physics-based baselines to state-of-the-art deep learning methods. %We describe the methods and experiments, % (described in Sec.~\ref{sec:experiments:methods}). We detail the experiment setup in Sec.~\ref{sec:experiments:setup} and present the results in Sec.~\ref{sec:experiments:results}.

\subsection{Prediction methods}
\label{sec:experiments:methods}

Our benchmark comes with several model- and learning-based methods \cite{helbing1995social,karamouzas2009predictive,zanlungo2011social,Gupta2018SocialGAN,salzmann2020trajectronpp}.
The Social force model \cite{helbing1995social} ({\bf Sof}) is a simple and well-known approach to interaction modeling of people in groups, used in applications such as crowd behavior analysis, simulation and animation \cite{yang2014guided, farina2017walking}, robotics \cite{luber2010} and human motion prediction \cite{Rudenko2018icra}. %Often a reasonable choice due to its reliability and simple implementation, the social force model suffers from inherent reactivity: the agents engage in passive collision avoidance only when in close proximity for the social forces to take effect (see Fig.~\ref{testing}). In reality, people adapt their trajectories to avoid collisions in advance. To correct this sort of behavior, the social force theory was extended with explicit collision prediction by a number of authors.
We also consider the extension of the social force model by Karamouzas et al. \cite{karamouzas2009predictive} ({\bf Kara}) who added a predictive ability to the initial approach  by forward projection of the agent's current motion and avoiding collisions in advance. The constant velocity motion model ({\bf CVM}) further serves as baseline in our experiments.

The shift towards learning-based methods for human motion prediction has produced a large number of new methods and models in recent years. For our example comparison, we 
%In the recent years the research focus in trajectory prediction shifted towards the deep learning approaches, which promise to learn and predict the interactions better than the model-based ones. In our comparison
include two state-of-the-art methods, Trajectron++  \cite{salzmann2020trajectronpp} ({\bf T++}), a graph-structured generative neural network based on a conditional-variational autoencoder and Social GAN \cite{Gupta2018SocialGAN} ({\bf SGAN}), which combines a recurrent sequence-to-sequence model with a generative adversarial network.

\subsection{Setup}
\label{sec:experiments:setup}

% The sheer breadth of experimentation scope study of motion prediction algorithms can be overwhelming.

In our evaluation we vary the testing parameters around the commonly used values of $O_s= 3.2$ and $T_s= 4.8$ \SI{}{\second}. As all datasets in our experiments are downsampled to \SI{2.5}{\Hz}, this implies $O_p=8$ and $T_p=12$. This is the standard setup in all experiments in  ATC and TH\"OR, where one parameter (for instance $T_p$, $O_p$, the amount of added noise $\sigma$, calibration/validation datasets) is varied. Due to the limited number of data in ETH, we set $O_s= 2.4$ and $T_s= 4.0$ (equivalent to $O_p=6$ and $T_p=10$) instead. To stress the agent interaction aspect of motion prediction, scenarios with less than 2 people are excluded from the evaluation. We report the mean and standard deviation of the ADE and FDE metrics across all scenarios in the experiment.
%\todo[inline]{LP: there is some confusion in the usage of O and T. Sometimes you use them for indicating seconds, sometimes as steps counters. We need to be consistent. Perhaps the ones reported in seconds can have a different naming, e.g. Os, Ts}

The force-based methods (Sof and Kara) are optimized separately in each dataset on the initial 30\% of the detections (20\% in case of ETH). The target optimization metric is FDE at $T_s=4.8$ \SI{}{\second}. For the optimization parameters of each individual method, we refer the reader to the provided implementations. The current velocity $\hat v^i_0$ of each person $i$, used as input to the force-based methods and the CVM, is calculated as a weighted sum of the finite differences in the observed trajectories. The sequence of past velocities $(v^i_{-1},\dots,v^i_{-O_p})$ is weighted with a zero-mean Gaussian filter with $\sigma=1.5$ to put more weight on the more recent observations: $\hat v^i_0 = \sum_{t=1}^{O_p} v^i_{-t} g(t)$, where $g(t) = \frac{1}{\sigma \sqrt{2\pi}} e^{-\frac{1}{2}(\frac{t}{\sigma})^2}$. The goal of each person is set to the point reached by forward propagating 40 steps into the future with $\hat v^i_0$.

The Trajectron++ implementation is accessed from the official repository\footnote{\url{https://github.com/StanfordASL/Trajectron-plus-plus}}, for which we provide a lightweight interface. This predictor is trained on the ETH dataset. We sample Trajectron++ once to get the most likely predicted trajectory.

For the Social GAN we use the implementation provided by the authors\footnote{\url{https://github.com/agrimgupta92/sgan}}. The model is trained on the ETH dataset, and is limited to accepting 8 frames as observations and producing 12 frames of prediction. Therefore, we exclude it from all experiments with $O_p < 8$ and $T_p > 12$. Similarly to T++, we sample SGAN for one mode.

\begin{table}[t!]
	\begin{center}
		\scriptsize{
			\begin{tabular}{|c|c|c|c|c|c|}
				\hline
				 &\multicolumn{5}{c|}{Prediction horizon}\\
				\cline{2-6} 
				& Method & 1.6 s & 3.2 s & 4.8 s & 8 s \\
				\hline
				\parbox[t]{0.5mm}{\multirow{5}{*}{\rotatebox[origin=c]{90}{ADE}}} 
				& CVM & $0.155 \pm.04$ & $0.319 \pm.09$ & $0.499 \pm.15$ & $0.870 \pm.30$ \\
				\cline{2-6} 
				& Sof & $0.156 \pm.05$ & $0.318 \pm.09$ & $0.494 \pm.15$ & $0.870 \pm.30$ \\
				\cline{2-6} 
				& Kara & $ 0.164 \pm.05$ & $0.324 \pm.09$ & $0.508 \pm.15$ & $ 0.872 \pm.31$ \\
				\cline{2-6} 
				& SGAN & $0.240 \pm .08$ & $0.500 \pm .15$ & $0.785 \pm .24$ & -- \\
				\cline{2-6} 
				& T++ & $0.152\pm.04$ & $ 0.340 \pm.09$ & $0.549 \pm.16$ & $1.006 \pm.29$ \\
				\hline \hline
				\parbox[t]{0.5mm}{\multirow{5}{*}{\rotatebox[origin=c]{90}{FDE}}} 
				& CVM & $0.272 \pm.08$ & $0.621 \pm.19$ & $1.000 \pm.33$ & $1.845 \pm.71$ \\
				\cline{2-6} 
				& Sof & $0.272 \pm.08$ & $0.620 \pm.19$ & $1.008 \pm.33$ & $1.846 \pm.71$ \\
				\cline{2-6} 
				& Kara & $0.279 \pm.08$ & $0.623 \pm.19$ & $1.000 \pm.33$ & $ 1.845 \pm.71 $ \\
				\cline{2-6} 
				& SGAN & $0.418 \pm .13$ & $0.977 \pm .31$ & $ 1.592 \pm .51$ & -- \\
				\cline{2-6} 
				& T++ & $0.273 \pm.08$ & $0.689 \pm.20 $ & $1.149 \pm.35 $ & $2.187 \pm.70 $ \\
				\hline
			\end{tabular}
		}
	\end{center}
	%\vspace{-10pt}
	\caption{ADE in the ATC dataset varying the prediction horizons}
	%\vspace{-10pt}
	\label{tab:atlas:pre_step_atc_ade}
\end{table}
\begin{table}[t!]
	\begin{center}
		\scriptsize{
			\begin{tabular}{|c|c|c|c|c|c|}
				\hline
				 &\multicolumn{5}{c|}{Prediction horizon}\\
				\cline{2-6} 
				& Method & 1.6 s & 3.2 s & 4.8 s & 8 s \\
				\hline
				\parbox[t]{0.5mm}{\multirow{5}{*}{\rotatebox[origin=c]{90}{ADE}}} 
				& CVM & $0.20 \pm0.09$ & $0.50 \pm0.22$ & $0.87 \pm0.39$ & $1.80 \pm0.73$ \\
				\cline{2-6} 
				& Sof & $0.29 \pm0.13$ & $0.54 \pm0.22$ & $0.82 \pm0.34$ & $1.42 \pm0.52$ \\
				\cline{2-6} 
				& Kara & $ 0.32 \pm0.14$ & $0.57 \pm0.23$ & $0.85 \pm0.35$ & $ 1.44 \pm0.54$ \\
				\cline{2-6} 
				& SGAN & $ 0.29 \pm 0.11$ & $ 0.65 \pm 0.22$ & $1.08 \pm 0.36$ & -- \\
				\cline{2-6} 
				& T++ & $0.18\pm0.07$ & $ 0.47 \pm0.18$ & $0.84 \pm0.33$ & $1.68 \pm0.62$ \\
				\hline \hline
				\parbox[t]{0.5mm}{\multirow{5}{*}{\rotatebox[origin=c]{90}{FDE}}} 
				& CVM & $0.38 \pm0.17$ & $1.07 \pm0.48$ & $1.95 \pm0.86$ & $4.10 \pm1.51$ \\
				\cline{2-6} 
				& Sof & $0.46 \pm0.20$ & $1.02 \pm0.44$ & $1.66 \pm0.74$ & $2.95 \pm1.08$ \\
				\cline{2-6} 
				& Kara & $0.49 \pm0.21$ & $1.05 \pm0.45$ & $1.69 \pm0.75$ & $ 2.96 \pm1.08 $ \\
				\cline{2-6} 
				& SGAN & $ 0.52 \pm 0.19$ & $1.36 \pm 0.47$ & $2.33 \pm 0.27$ & -- \\
				\cline{2-6} 
				& T++ & $0.34 \pm0.14$ & $1.04 \pm0.42 $ & $1.91 \pm0.75 $ & $3.79 \pm1.28 $ \\
				\hline
			\end{tabular}
		}
	\end{center}
	%\vspace{-10pt}
	\caption{ADE in the TH\"OR3 dataset varying the prediction horizon}
	%\vspace{-10pt}
	\label{tab:atlas:pre_step_thor3_ade}
\end{table}

In addition to the accuracy measurements under various conditions, we also estimate the prediction runtime conditioned on the number of people in the scenario. All experiments were executed on a laptop with an Intel i7 2.7 GHz 12 core processor and 32 GB of RAM.

Note that the evaluation results may differ from the original papers \cite{alahi2016social,Gupta2018SocialGAN, salzmann2020trajectronpp}, due to the differences in the evaluation protocols, which are not fully disclosed and therefore not reproducible. This particularly highlights the need for standardized benchmarks like Atlas, which allow for transparent verification of reported results and direct comparisons of methods under various conditions of interest.

\subsection{Results and Discussion}
\label{sec:experiments:results}

\begin{figure}[t!]
	\centering
	\includegraphics[width=0.49\textwidth]{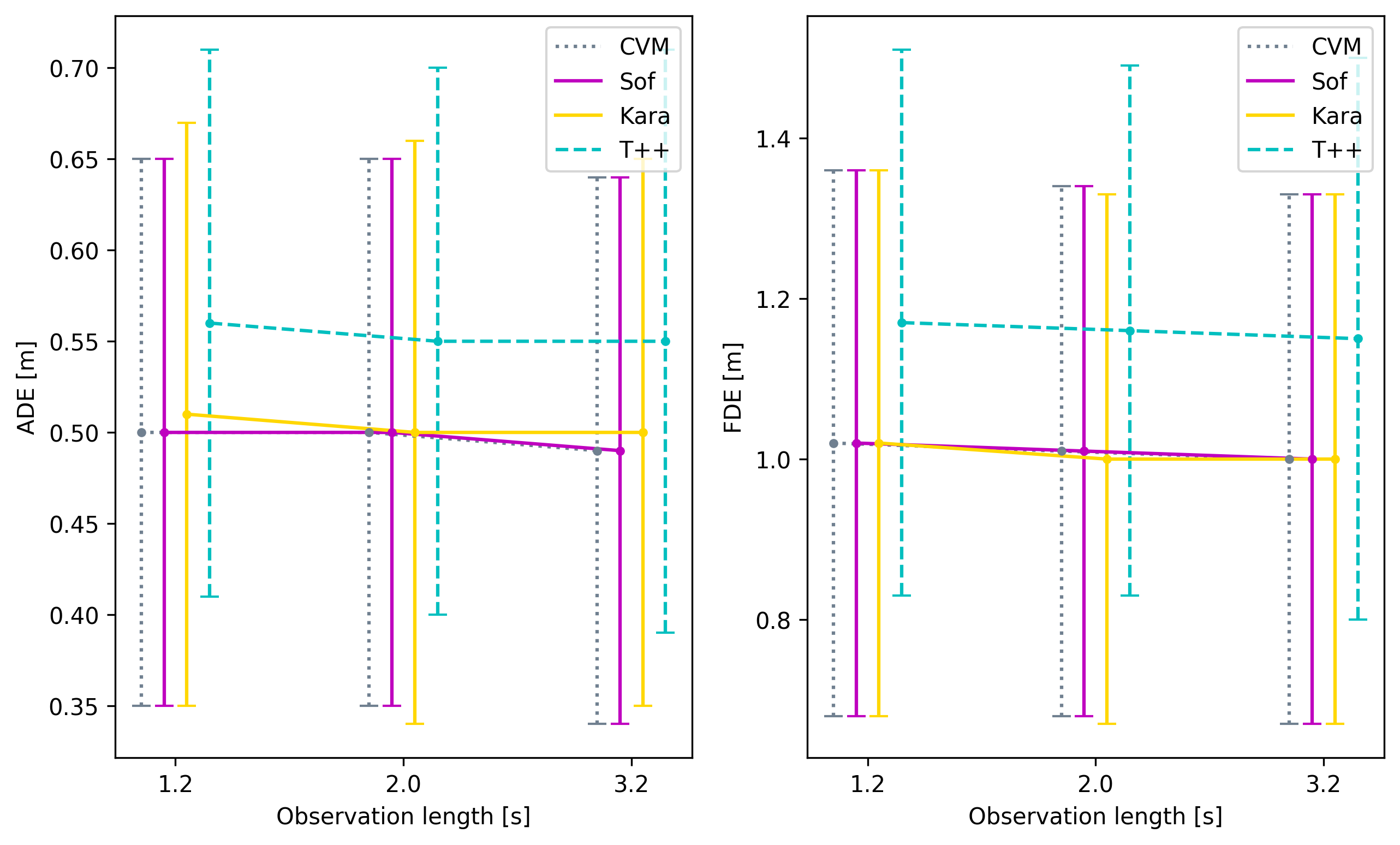}
	%\vspace{-18pt}
	\caption[Observation length variation experiment in the ATC dataset]{ADE/FDE in the ATC dataset with different observation lengths}
	%\vspace{-5pt}
	\label{fig:atlas:obs_atc_nsg}
\end{figure}

\begin{figure}[t!]
	\centering
	\includegraphics[width=0.49\textwidth]{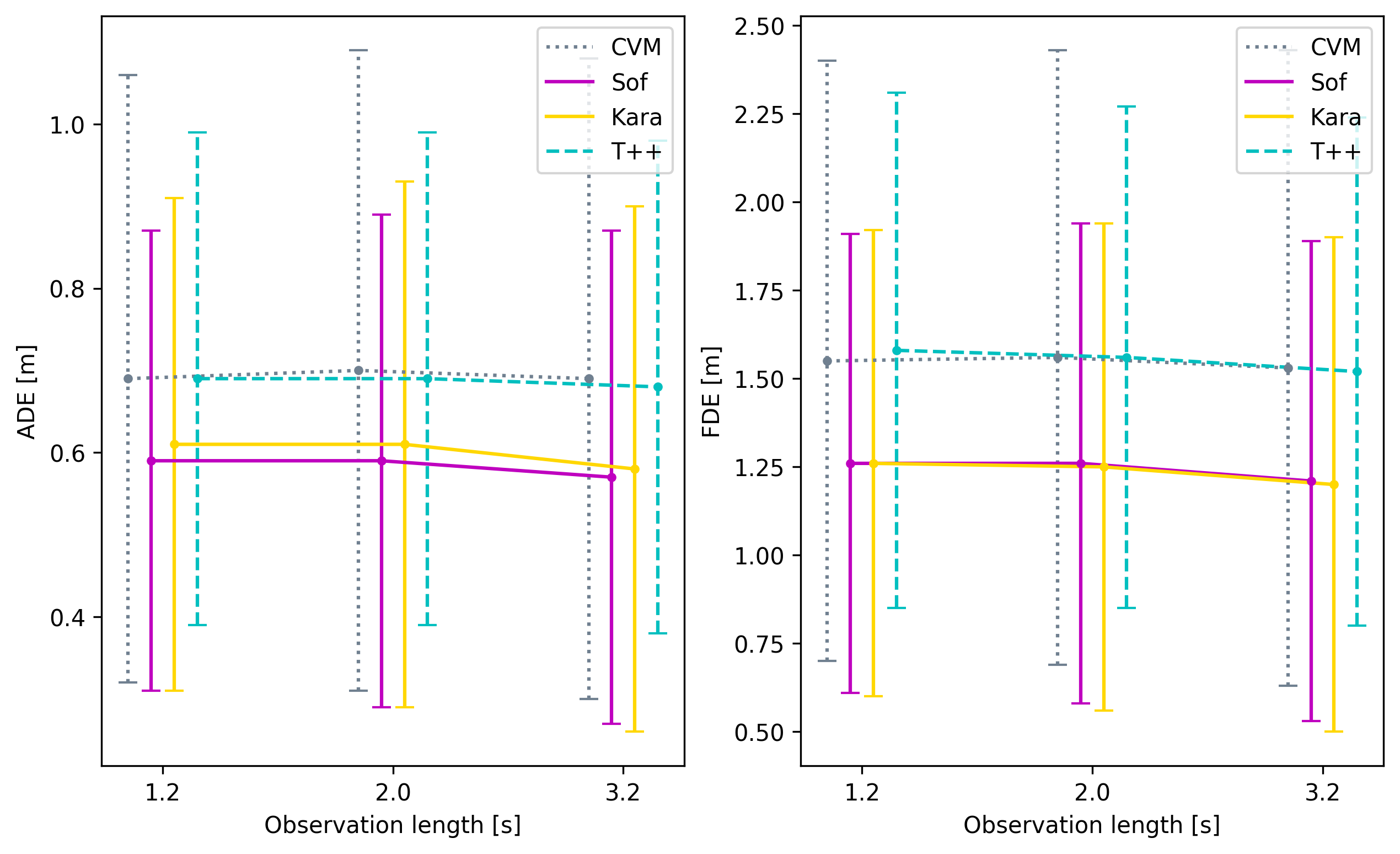}
	%\vspace{-18pt}
	\caption{ADE/FDE in the TH\"OR1 dataset with different observation lengths}
	\label{fig:atlas:obs_thor}
\end{figure}

\begin{figure}[t]
	\centering
	\includegraphics[width=0.49\textwidth]{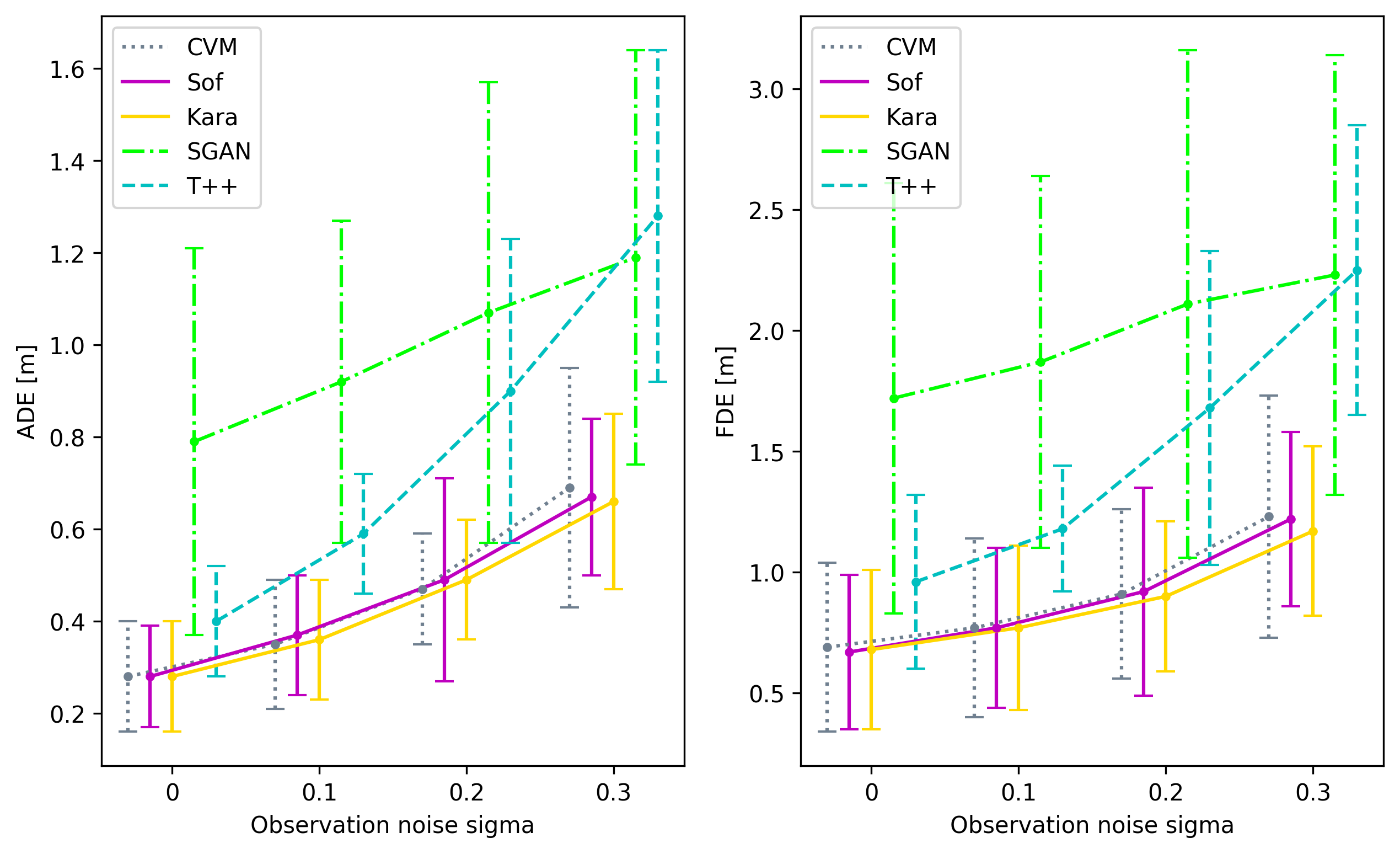}
	%\vspace{-18pt}
	\caption{ADE/FDE in the ETH dataset with added noise}
	\label{fig:atlas:robustness_eth}
\end{figure}

\begin{figure}[t]
	\centering
	\includegraphics[width=0.49\textwidth]{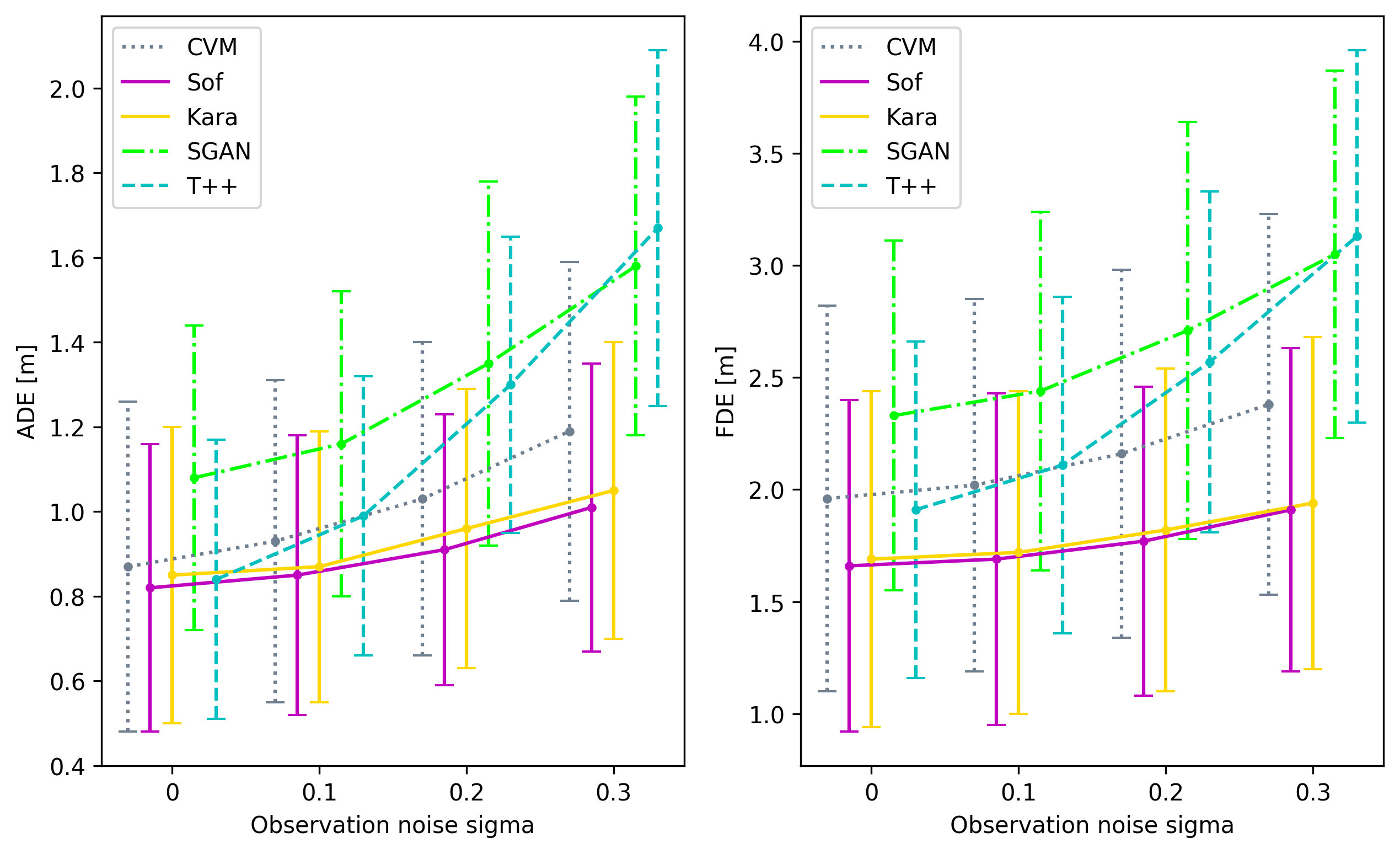}
	%\vspace{-18pt}
	\caption{ADE/FDE in the TH\"OR3 dataset with added noise} 
	\label{fig:atlas:robustness_thor3}
\end{figure}

\begin{figure}[t]
	\centering
	\includegraphics[width=0.237\textwidth]{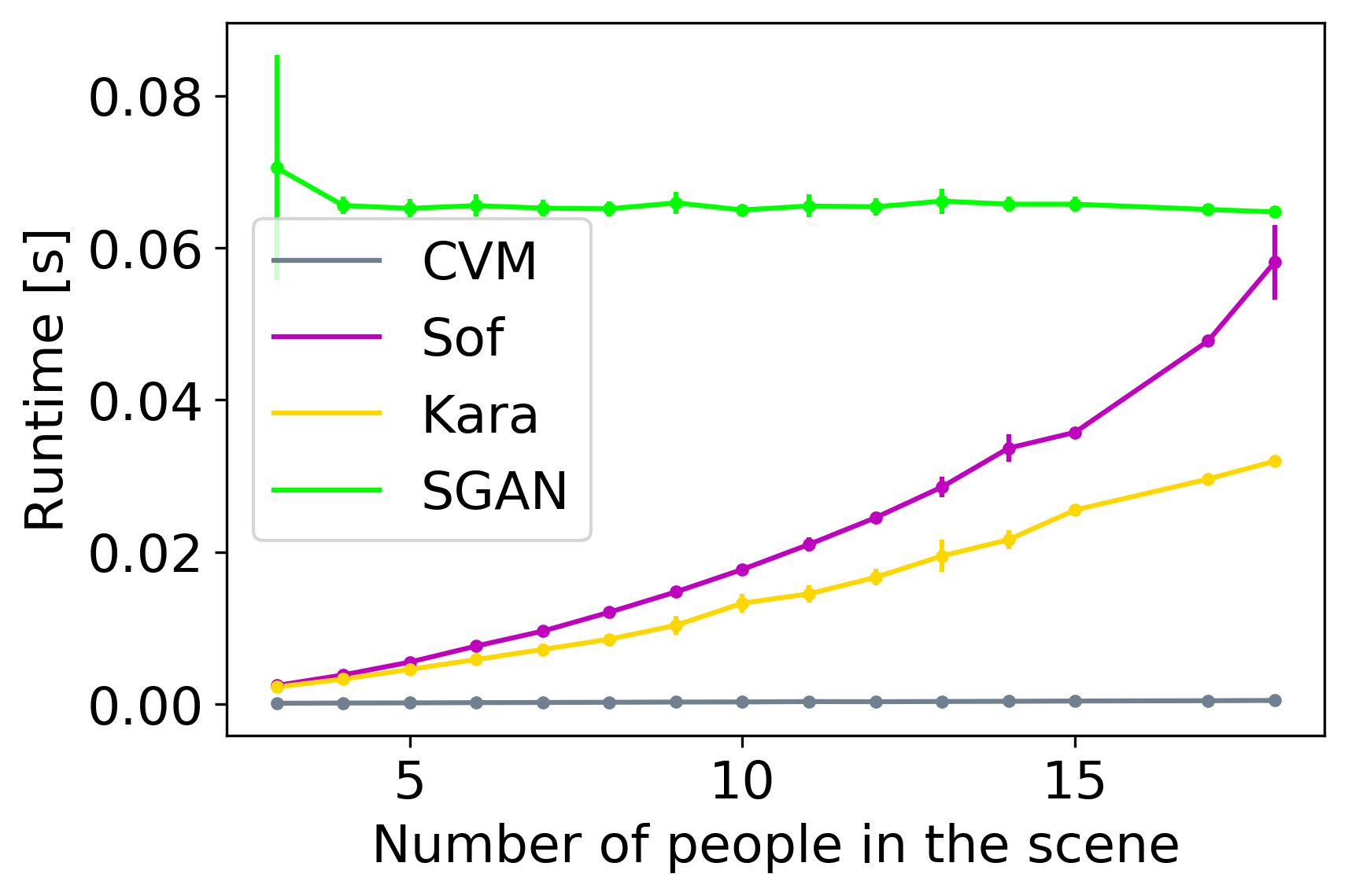}
	\hspace{0.171cm}
	\includegraphics[width=0.22\textwidth]{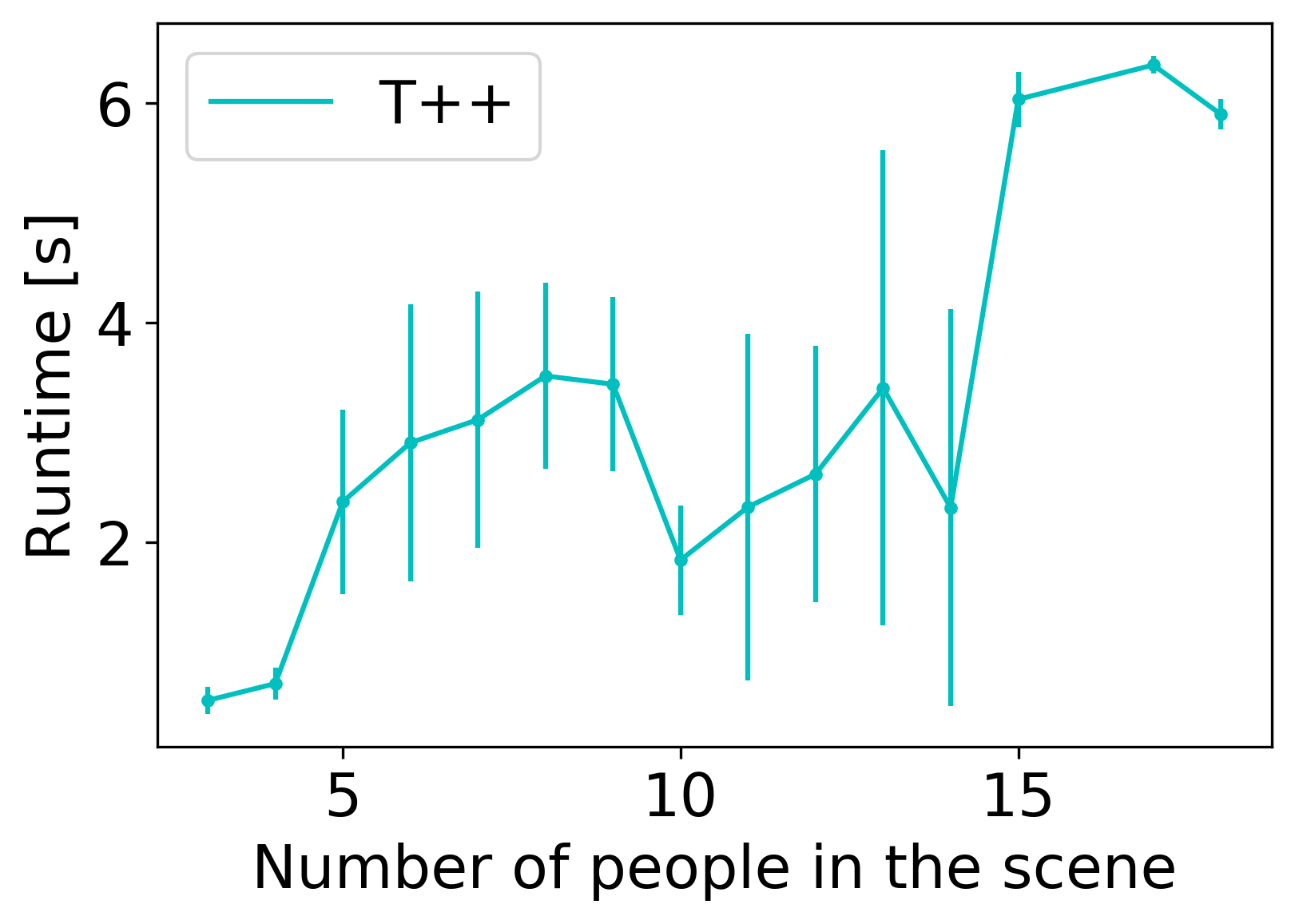}
	%\vspace{-8pt}
	\caption{Average runtimes to compute predictions for $O_s=3.2$ \SI{}{\second} and $T_s=4.8$ \SI{}{\second} in the scenes from the ATC dataset, sorted by the number of people. {\bf Left}: model-based methods and SGAN, {\bf right}: Trajectron++. Despite achieving roughly comparable performance, the model-based methods are two orders of magnitude faster than the Trajectron++. SGAN has constant runtime performance. The irregular shape of the Trajectron++ performance curve is explained by the number of and the irregularities in the scenarios: due to the pooling and pruning when computing interactions, a scene with 10 people far away from each other might be easier to solve than with 6 people closely interacting.}
	\label{fig:atlas:runtime}
\end{figure}

\begin{figure*}[t!]
	\centering
	\includegraphics[width=0.342\textwidth]{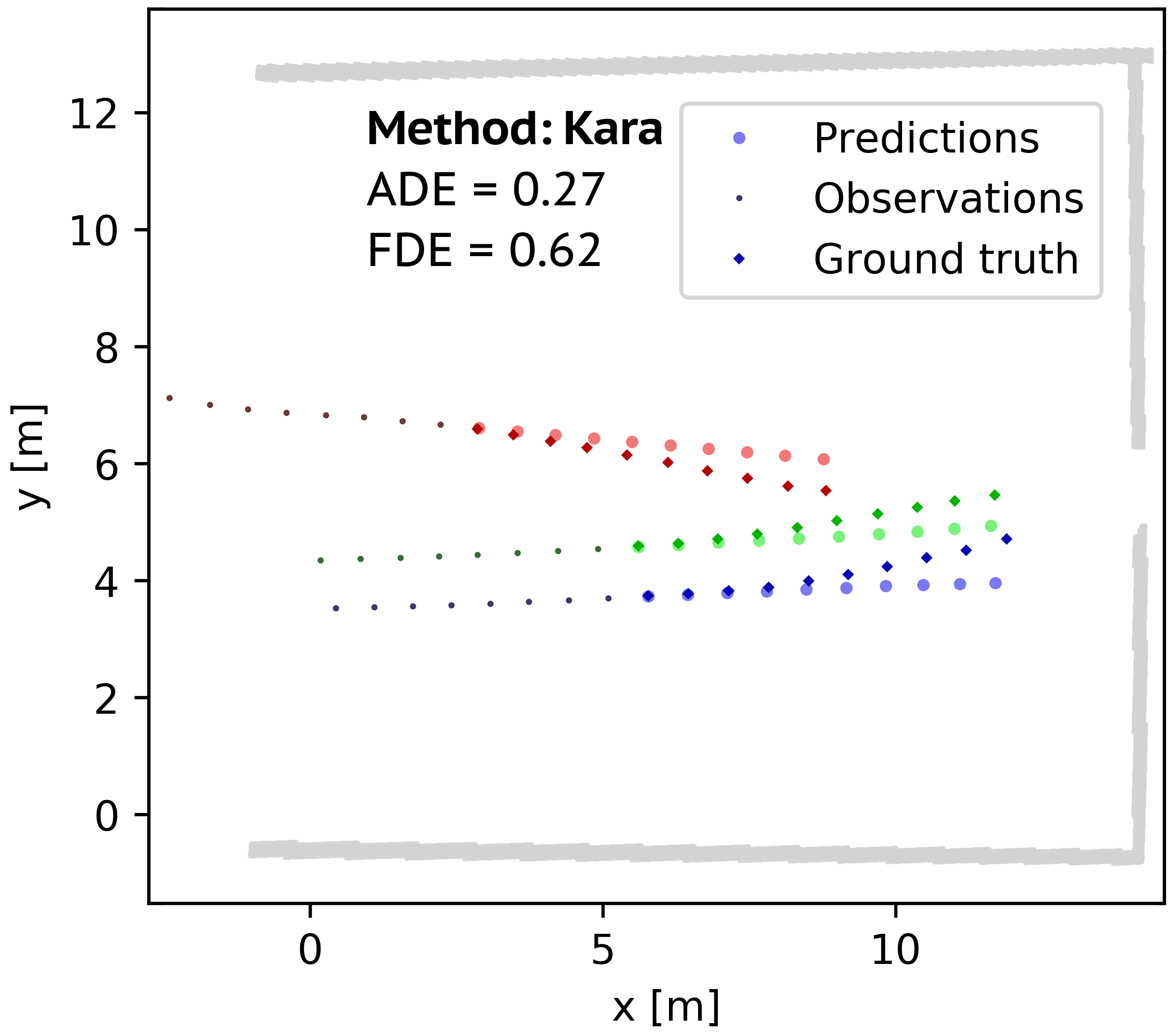}
	\includegraphics[width=0.302\textwidth]{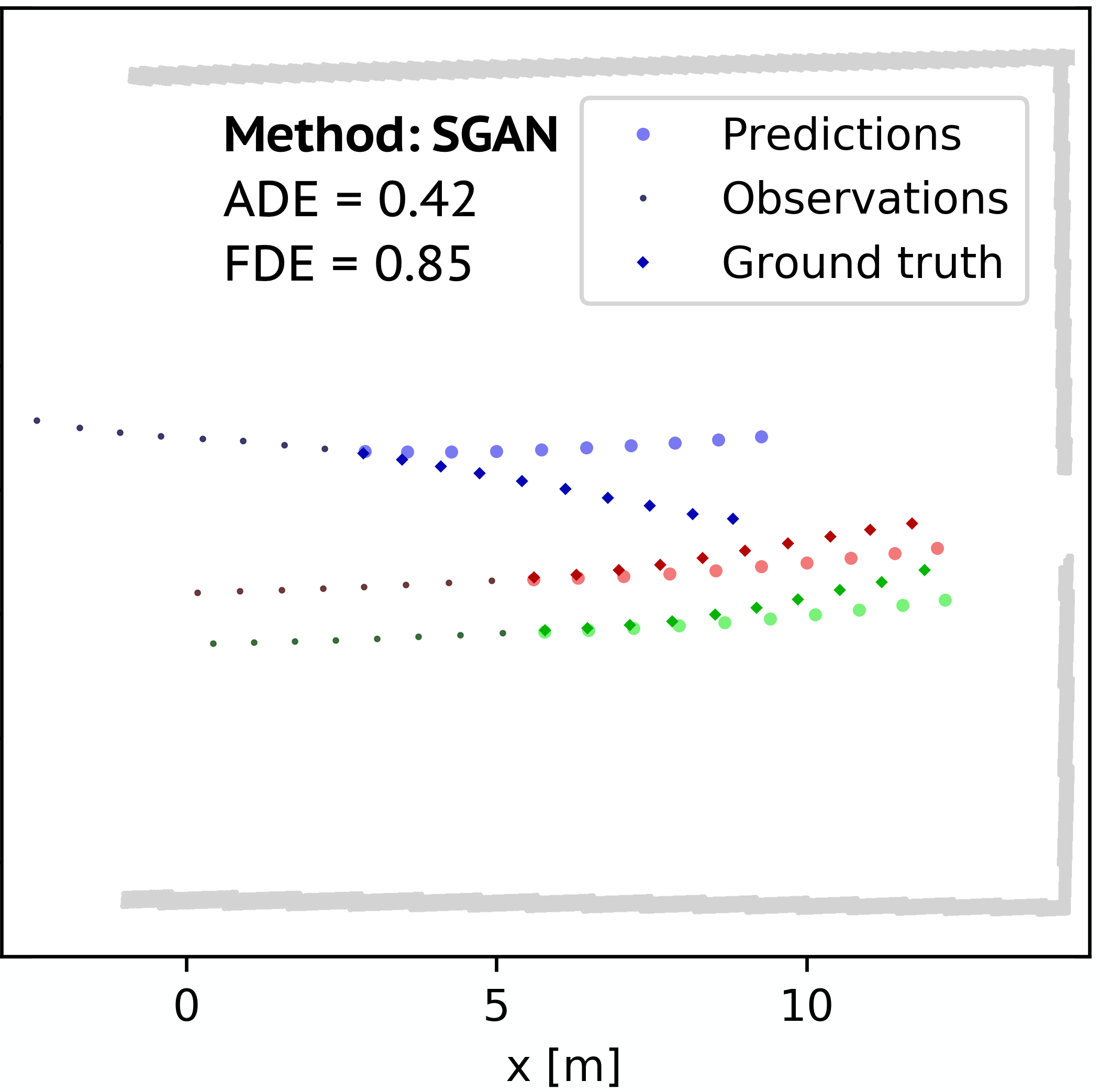}
	\includegraphics[width=0.30\textwidth]{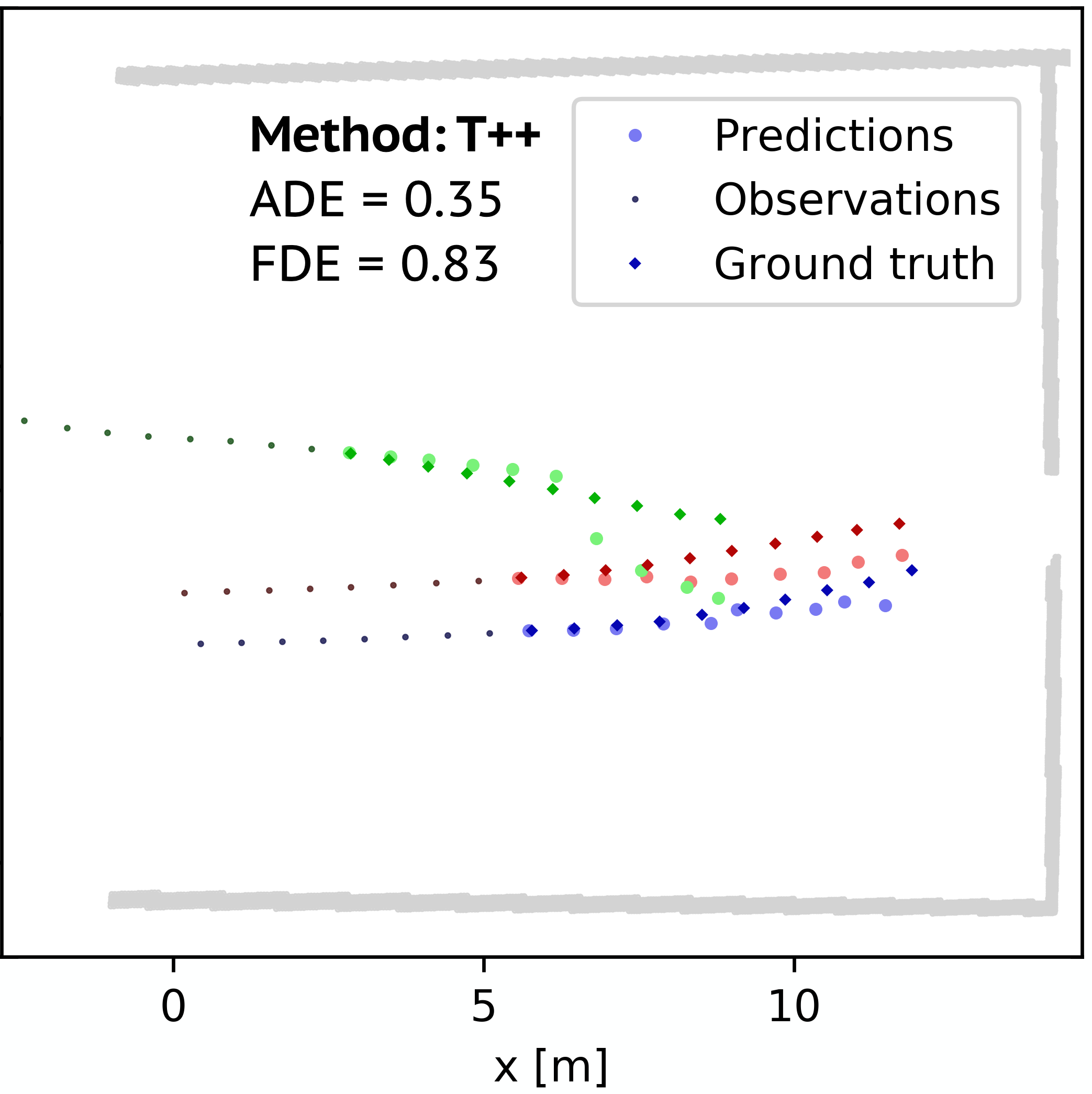}
	\vspace{-8pt}
	\caption{Predictions in the ETH scenario} 
	\label{fig:atlas:predictions:eth}
\end{figure*}

\begin{figure*}[t!]
	\centering
	\includegraphics[width=0.356\textwidth]{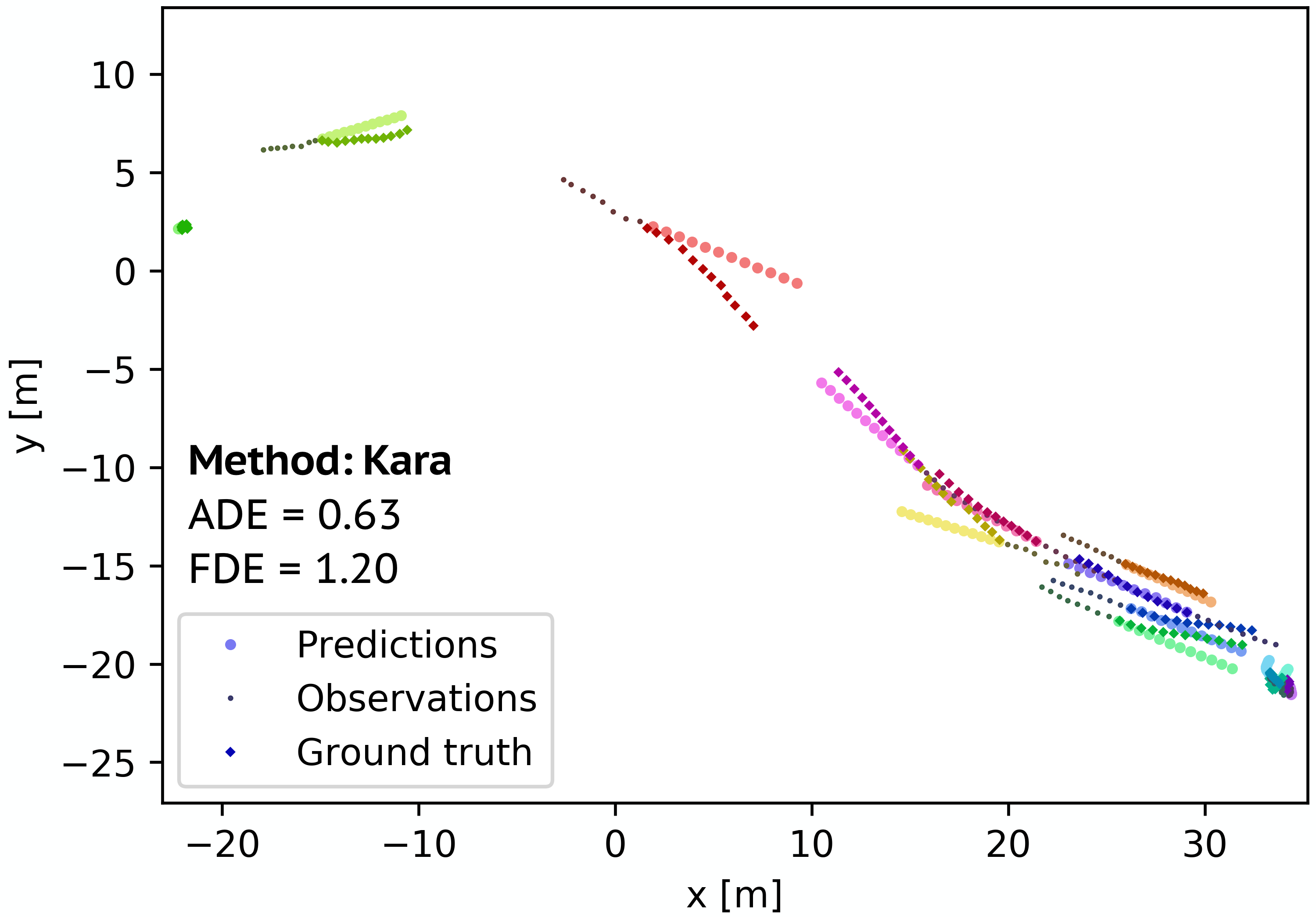}
	\includegraphics[width=0.312\textwidth]{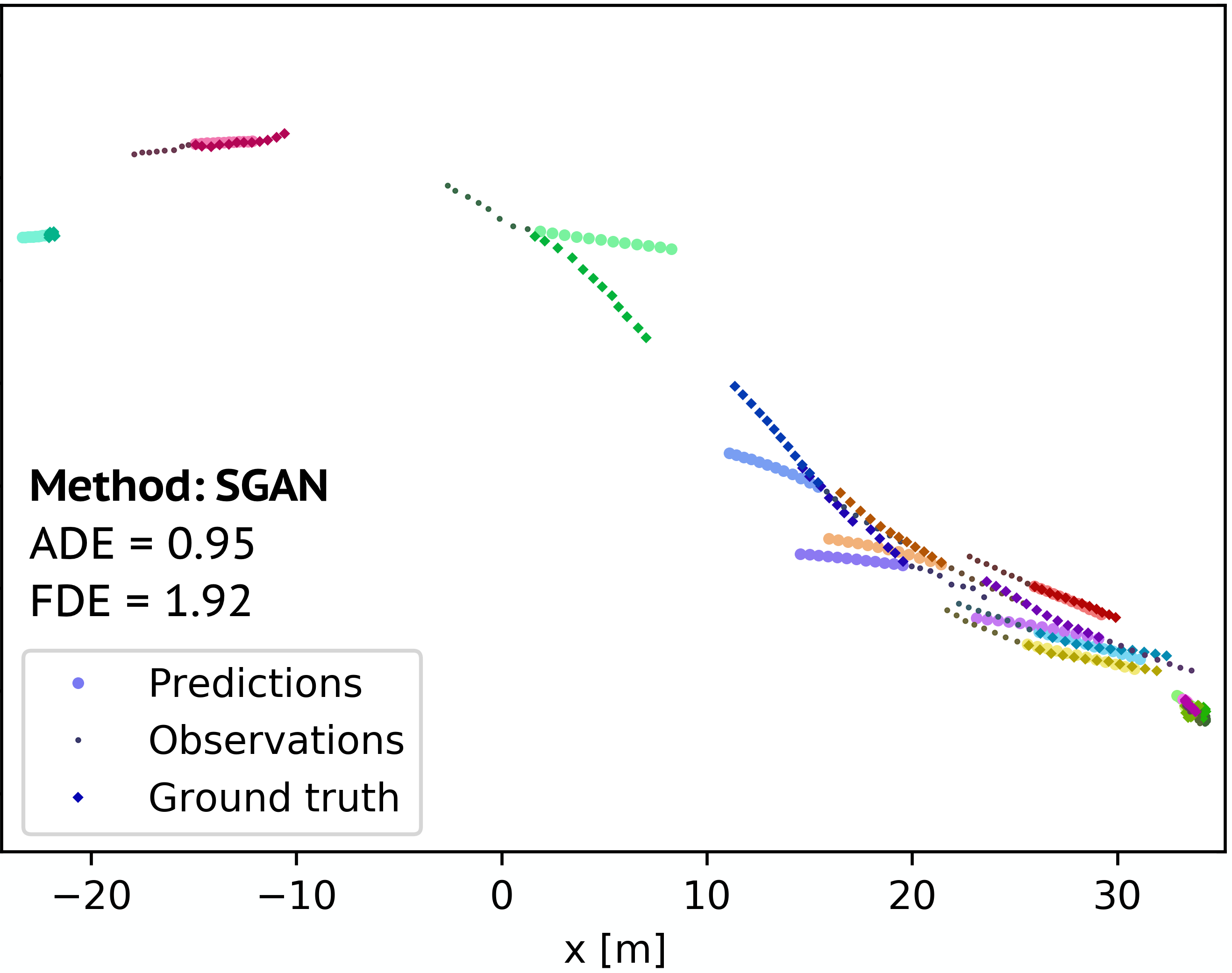}
	\includegraphics[width=0.312\textwidth]{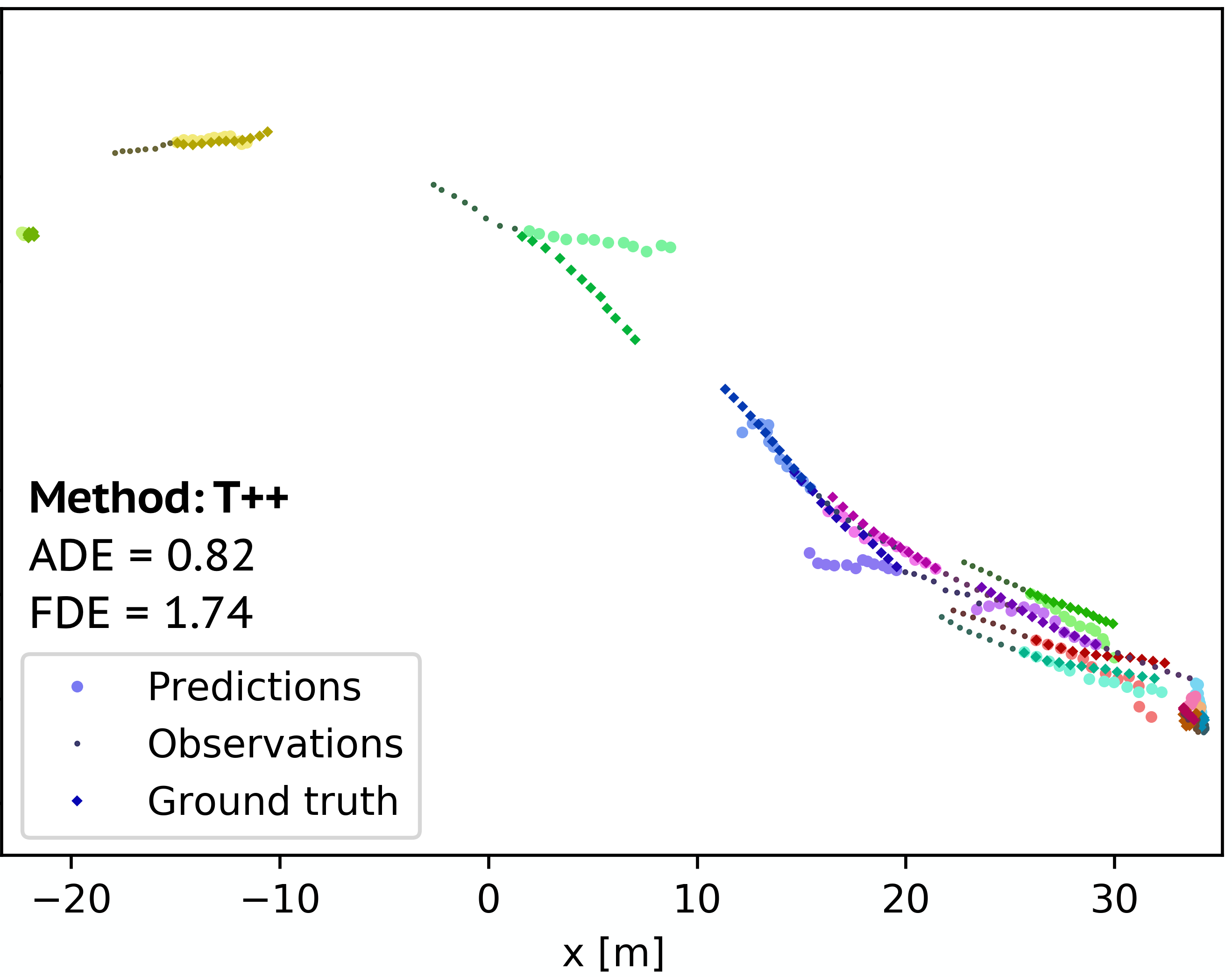}
	\vspace{-8pt}
	\caption{Predictions in the ATC scenario} 
	\label{fig:atlas:predictions:atc}
\end{figure*}

\begin{figure*}[t!]
	\centering
	\includegraphics[width=0.277\textwidth]{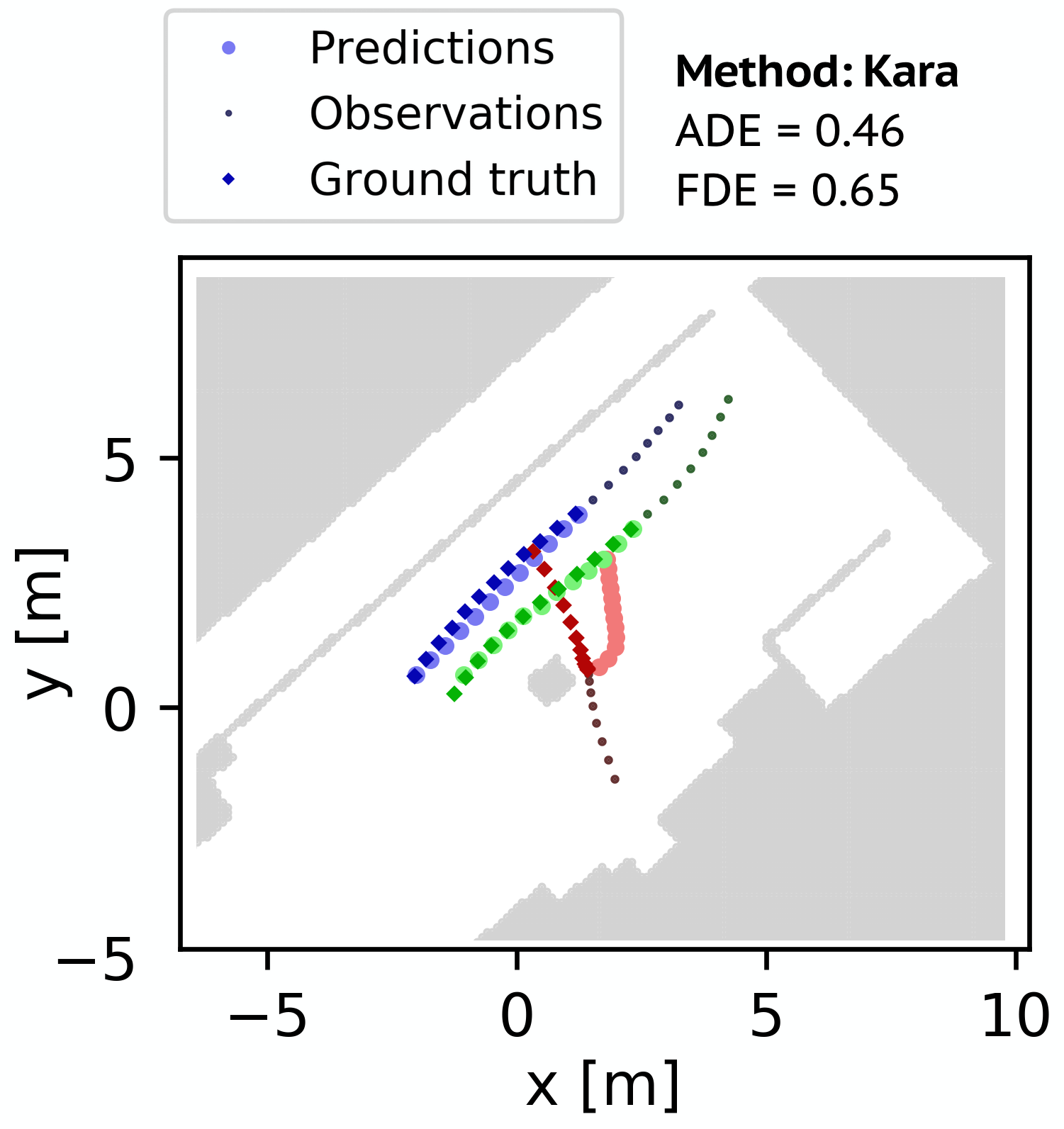}
	\includegraphics[width=0.27\textwidth]{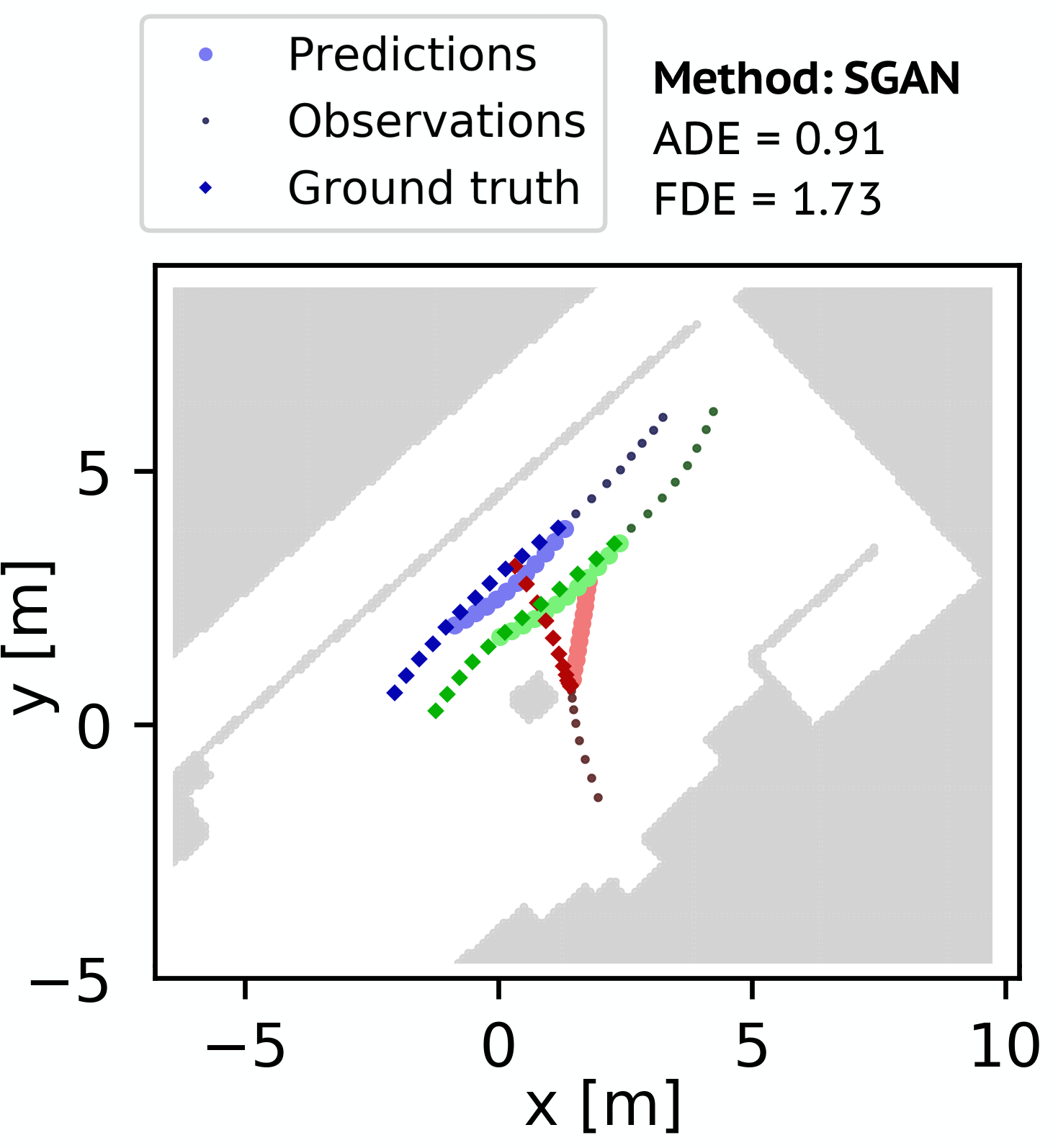}
	\includegraphics[width=0.277\textwidth]{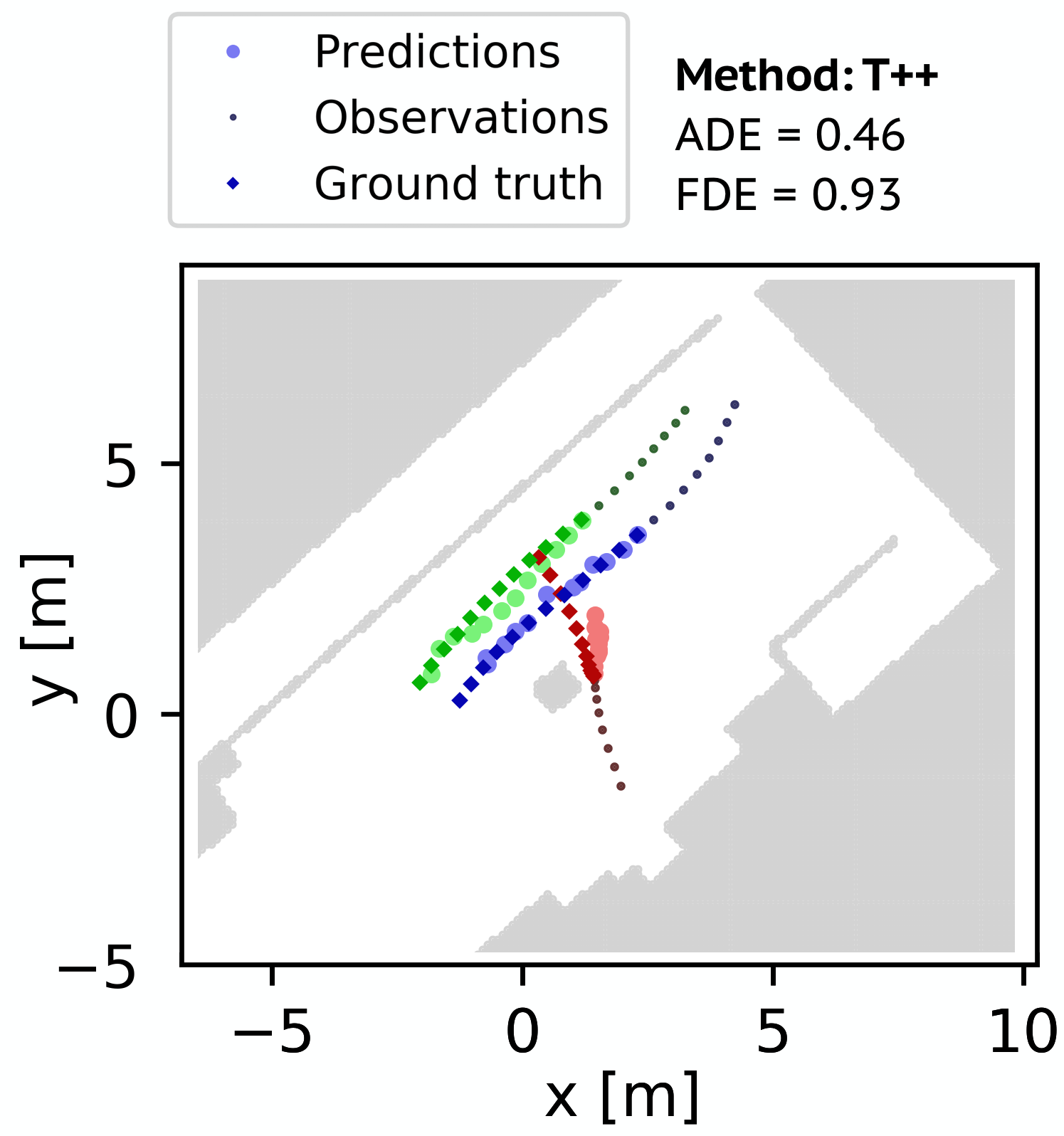}
	\vspace{-8pt}
	\caption{Predictions in the TH\"OR1 scenario} 
	\label{fig:atlas:predictions:thor1}
\end{figure*}

\begin{figure*}[t!]
	\centering
	\includegraphics[width=0.272\textwidth]{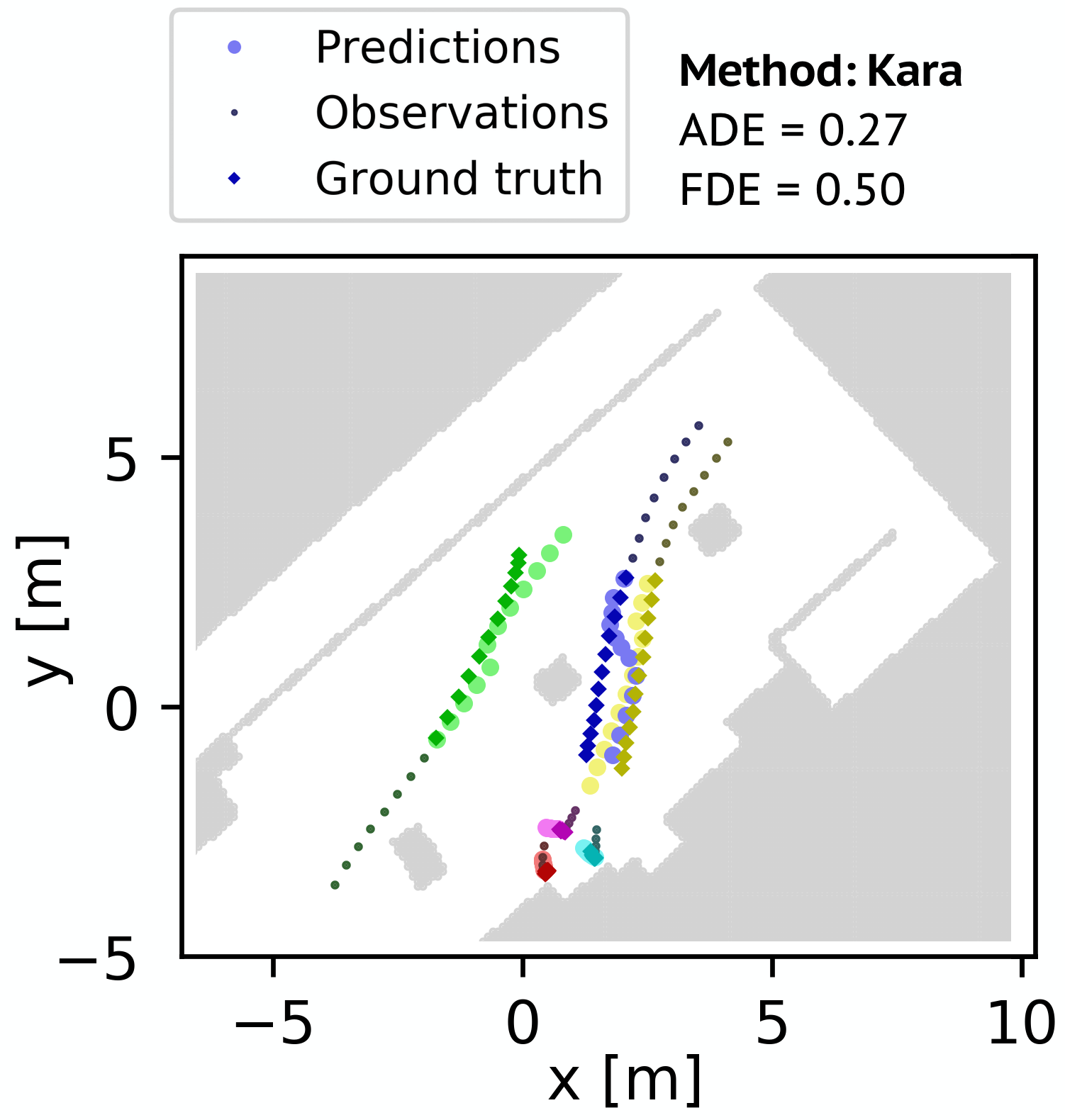}
	\includegraphics[width=0.273\textwidth]{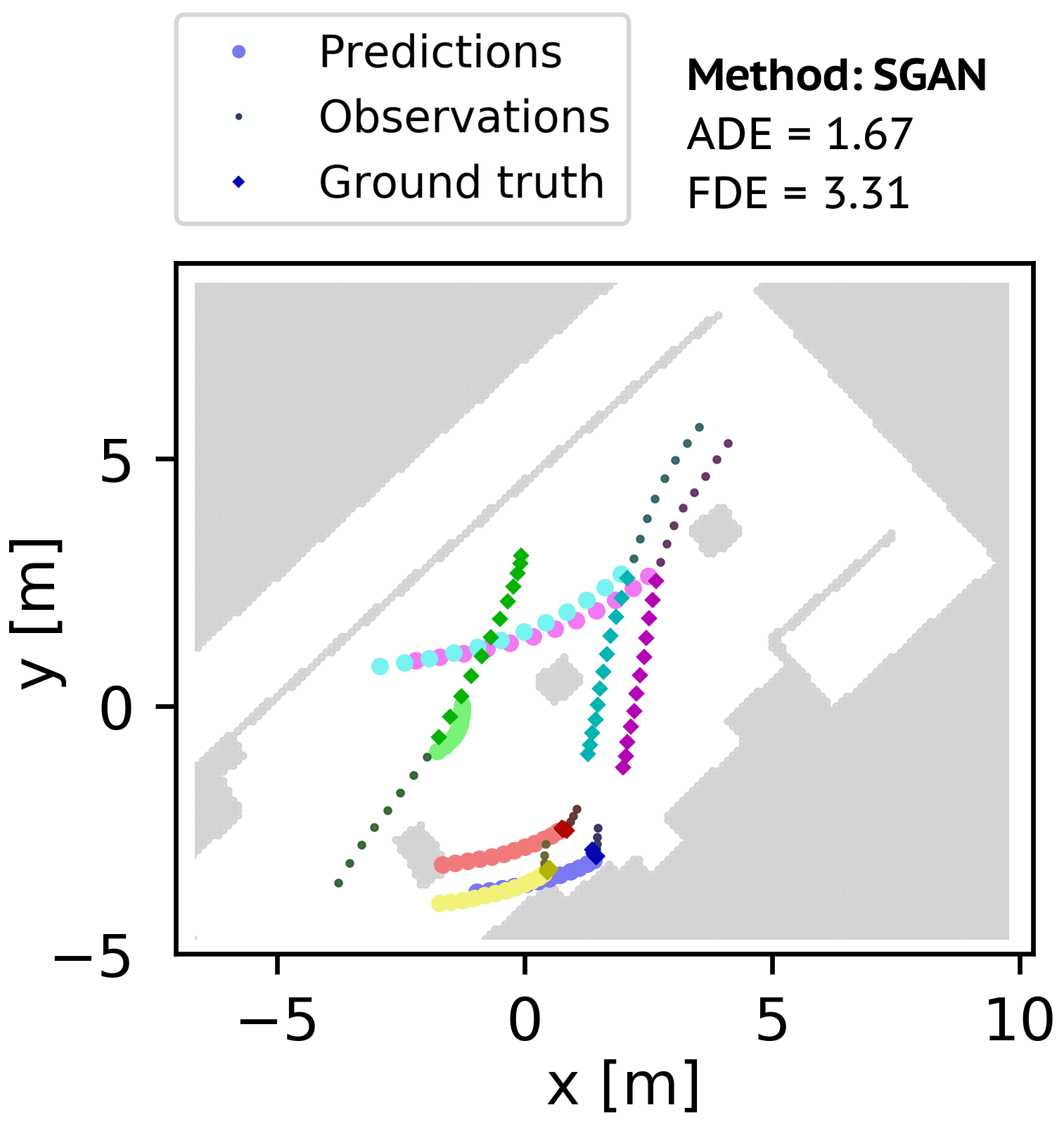}
	\includegraphics[width=0.27\textwidth]{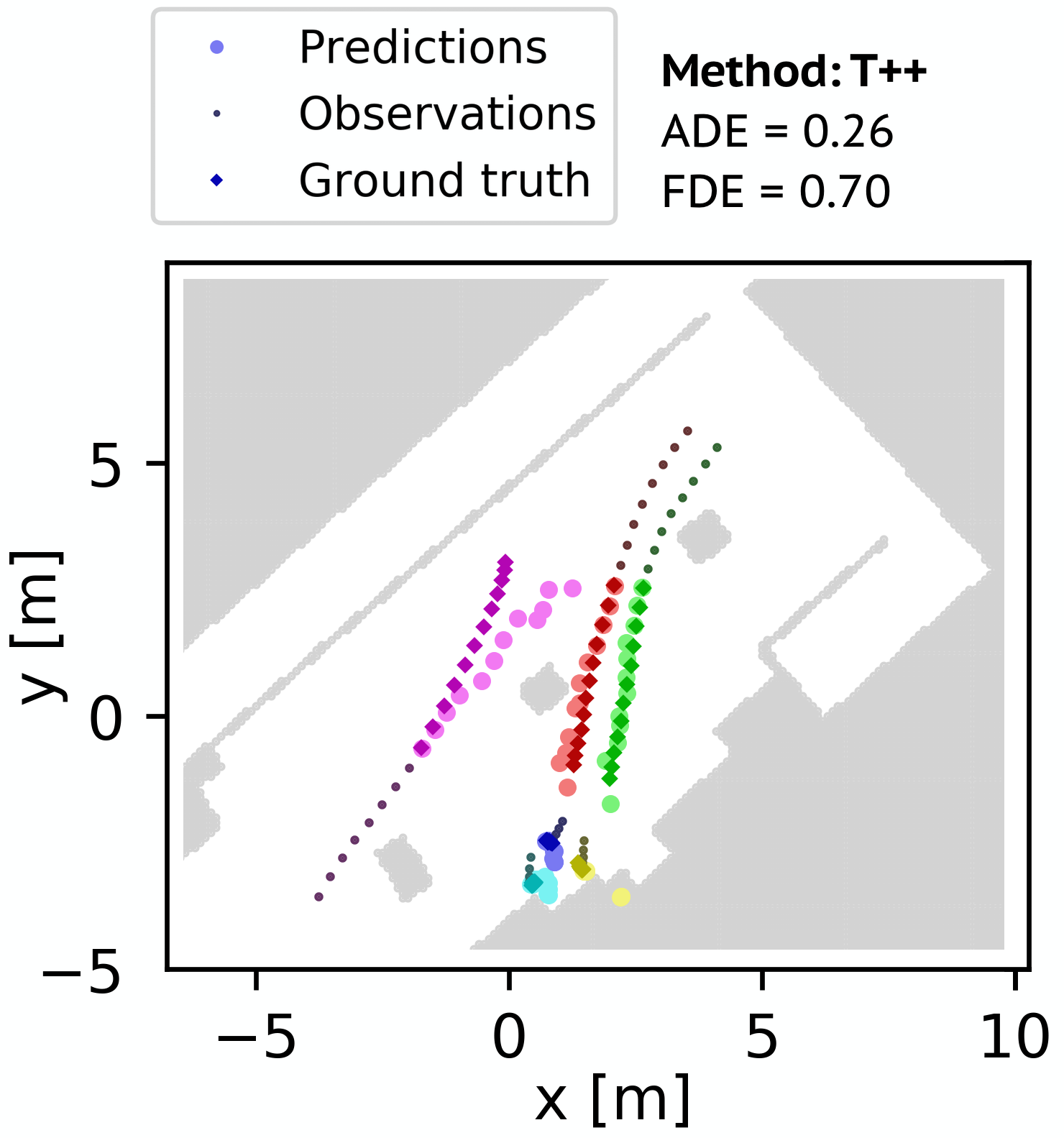}
	\vspace{-8pt}
	\caption{Predictions in the TH\"OR3 scenario} 
	\label{fig:atlas:predictions:thor3}
\end{figure*}

We present the results in Tables~\ref{tab:atlas:pre_step_atc_ade}--\ref{tab:atlas:transfer_exp_ade}, Fig.~\ref{fig:atlas:obs_atc_nsg}--\ref{fig:atlas:runtime} and show example predictions in Fig.~\ref{fig:atlas:predictions:eth}--\ref{fig:atlas:predictions:thor3}. Due to space limitations, results of each experiment are presented in selected datasets only, but the discussed trends are observed in all of them.

Tables~\ref{tab:atlas:pre_step_atc_ade} and \ref{tab:atlas:pre_step_thor3_ade} show the results of evaluating the ADE and FDE on different prediction horizons with the fixed observation period $O_s=3.2$ \SI{}{\second}. In the ATC dataset, which contains mostly straight linear motion, even in crowded scenes, the force-based approaches perform on the level of constant velocity. Trajectron++ and SGAN, on the other hand, attempt to predict more variety in motion than what exists in real life, leading to higher displacement errors (see an example scenario in Fig.~\ref{fig:atlas:predictions:atc}).

In the TH\"OR3 dataset (Table~\ref{tab:atlas:pre_step_thor3_ade} and Fig.~\ref{fig:atlas:predictions:thor3}), on the contrary, people navigate in a tighter environment across multiple directions, increasing the importance of good interaction modeling. Here Trajectron++ outperforms the CVM, however the best and most stable results are reached by the force-based methods.


In the experiments with different observation horizons we found all methods to perform very robustly even with  observation lengths as short as 1.2 \SI{}{\second}, see Fig.~\ref{fig:atlas:obs_atc_nsg}~and~\ref{fig:atlas:obs_thor}. The ADE/FDE results for increasing amounts of noise in the agent positions %($\sigma=0.1$)
added to the ETH and TH\"OR3 data are shown in
Fig.~\ref{fig:atlas:robustness_eth}~and~\ref{fig:atlas:robustness_thor3}. The performance of all methods, including Trajectron++, degrades considerably when the observations become more unreliable (with $\sigma \geq 0.2$) where SGAN shows an almost linear degradation to the amount of noise, as compared to exponential decrease of other methods. Again, the model-based methods outperform the learning-based ones.

\begin{table}[t]
	\begin{center}
		\scriptsize{
			\begin{tabular}{|p{0.02cm}|p{0.03cm}|p{2.00cm}|p{1.18cm}|p{1.18cm}|p{1.18cm}|}
				\hline
				\multicolumn{1}{|p{0.01cm}|}{} & \multicolumn{5}{c|}{Test}\\ 
				\hline
				&&&&&\\ [-1em]
				\parbox[t]{2mm}{\multirow{12}{*}{\rotatebox[origin=c]{90}{Calibrate}}}  &
				\multicolumn{1}{p{0.6cm}|}{Dataset} &
				\multicolumn{1}{c|}{ETH} &
				\multicolumn{1}{c|}{ATC} &
				\multicolumn{1}{c|}{TH\"OR1} &
				\multicolumn{1}{c|}{TH\"OR3} \\ 
				\cline{2-6} 
				& ETH   & \begin{tabular}[c]{@{}r@{}}CVM: $0.283 \pm.12$\\ Sof: $0.277 \pm.11$\\ Kara: $0.278 \pm.12$\\ SGAN: $0.787 \pm .42$\\ T++: $0.399 \pm.36$\end{tabular} 
				& \begin{tabular}[c]{@{}l@{}}$0.499 \pm.15$\\ $0.494 \pm.15$ \\ $0.498 \pm.15$\\ $0.785 \pm .24$\\ $0.549 \pm.16$\end{tabular}
				& \begin{tabular}[c]{@{}l@{}}$0.69 \pm.39$\\$0.58 \pm.32$\\ $0.62 \pm.33$\\ $0.94 \pm.39$\\ $0.66 \pm.30$\end{tabular} 
				& \begin{tabular}[c]{@{}l@{}}$0.87 \pm.39$\\ $0.80 \pm.34$ \\ $0.81 \pm.38$\\ $1.08 \pm.36$\\ $0.84 \pm.33$\end{tabular} 
				\\ \cline{2-6} 
				& ATC & \begin{tabular}[c]{@{}r@{}} Sof: $0.29 \pm.13$\\ Kara: $0.34 \pm.15$\end{tabular}                    
				& \begin{tabular}[c]{@{}l@{}}$0.497 \pm.15$\\ $0.501 \pm.16$ \end{tabular} 
				& \begin{tabular}[c]{@{}l@{}}$0.67 \pm.37$\\ $0.58 \pm.31$ \end{tabular} 
				& \begin{tabular}[c]{@{}l@{}}$0.85 \pm.38$\\ $0.81 \pm.36$ \end{tabular}
				\\ \cline{2-6} 
				& TH\"OR1   & \begin{tabular}[c]{@{}r@{}}Sof: $0.31 \pm.14$\\Kara: $0.28 \pm.11$\end{tabular}                     
				&  \begin{tabular}[c]{@{}l@{}}$0.491 \pm.15$\\ $0.493 \pm.15$ \end{tabular} 
				&  \begin{tabular}[c]{@{}l@{}}$0.57 \pm.30$\\ $0.58 \pm.32$ \end{tabular} 
				&  \begin{tabular}[c]{@{}l@{}}$0.76 \pm.34$\\ $0.81 \pm.34$ \end{tabular}\\ 
				\cline{2-6} 
				& TH\"OR3 & \begin{tabular}[c]{@{}r@{}}Sof: $0.29 \pm.12$\\ Kara: $.28 \pm.11$\end{tabular}                     
				& \begin{tabular}[c]{@{}l@{}}$0.50 \pm.15$\\ $0.49 \pm.15$ \end{tabular} 
				& \begin{tabular}[c]{@{}l@{}}$0.60 \pm.32$\\ $0.61 \pm.33$ \end{tabular} 
				&  \begin{tabular}[c]{@{}l@{}} $0.82 \pm.34$\\ $0.85 \pm.35$ \end{tabular}\\
				\cline{2-6} 
				\hline
			\end{tabular}
		}
	\end{center}
	%\vspace{-10pt}
	\caption{ADE in the transfer experiments on different datasets}
	%\vspace{-10pt}
	\label{tab:atlas:transfer_exp_ade}
\end{table}


Table \ref{tab:atlas:transfer_exp_ade} summarizes the transfer experiment, where the methods are calibrated on one dataset and tested on another. We observe that the predictive social force approach (Kara) delivers more stable transfer performance in all cases as compared to the Sof method.

An overall conclusion from the experiments, supported by the qualitative analysis in Fig.~\ref{fig:atlas:predictions:eth}--\ref{fig:atlas:predictions:thor3}, is that the model-based prediction methods, properly calibrated, with velocity filtering and goal projection, offer a surprisingly competitive alternative to the complex state-of-the-art deep learning approaches. This result seems to confirm the recent findings by Sch\"oller et al. \cite{scholler2019simpler}, once again indicating that learning interactions is an extremely challenging task prone to evaluation pitfalls. That, and the considerable runtime differences in favor of the model-based approaches in Fig.~\ref{fig:atlas:runtime}, justifies the need for further research into interaction models, both engineered and learned ones. Another conclusion is that, in our experiments, the predictive social force model does not reliably outperform the original method. Finally, the results of the force-based methods calibrated on simpler datasets with a lot of linear motion (such as the ETH and ATC) converge to the CVM model up to the 3rd decimal digit (i.e. less than 1 \SI{}{\cm} difference).

%\section{Conclusion}
\section{Summary and Future Work}
\label{sec:conclusion}
%%% Initial version, koa
%%%In this paper we present the Atlas benchmark for motion prediction algorithms which is designed for evaluation and comparison in automated reproducible experiments with a systematic variation of the key motion prediction parameters. In future work we intend to add more datasets of human motion and extend the support for novel motion cues, such as the group motion, articulated pose or gaze direction where available. We intend to add support for non-human dynamic obstacles such as the robots in the environment. We are also looking to add further baselines, in particular model-based multi-modal ones, to conduct experiments with motion uncertainty and meaningful multi-modality, e.g. goal-driven or induced by the obstacle topology.

%%% koa version
The number of approaches for human motion prediction has grown rapidly in recent years but different datasets and varying evaluation protocols make in-depth analysis and comparisons difficult. This is the motivation for Atlas, introduced in this paper, an evaluation benchmark for motion prediction that enables researchers to analyze and compare their methods in an unified easy-to-use framework. Unlike related benchmarks and challenges, Atlas offers data preprocessing functions, hyperparameter optimization, three popular datasets and the fexibility to setup and conduct underexplored yet relevant experiments to stress a method's accuracy and robustness. In an example application of Atlas, we compared five popular prediction methods, three early physics-based approaches and two learning-based state-of-the-art approaches and found that the model-based methods, properly applied, are surprisingly competitive. While these findings motivate further research particularly in agent interaction modeling, they also show the necessity for such benchmarks to reproduce, confirm and further extend such results.

In future work we intend to extend Atlas with support for more datasets, more baselines, more motion cues such as group motion or articulated body pose, non-human dynamic agents such as vehicles or robots and other relevant environment descriptors, e.g. maps of dynamics \cite{kucner2020probabilistic}. Furthermore, considering the downstream performance metrics in how the predictions affect the robot behavior, we seek to close the loop on the human motion prediction benchmarking by connecting it to robot navigation simulation \cite{heiden2021bench}.

%\bibliographystyle{unsrt}
%\bibliography{bibliography_full}
\bibliographystyle{IEEEtran}
\bibliography{IEEEabrv,rudenkoHuang2021RSSws}

\end{document}